\documentclass[10pt,twocolumn,letterpaper]{article}

\usepackage{cvpr}              
\usepackage{graphicx}
\usepackage{amsmath}
\usepackage{amssymb}
\usepackage{amsfonts}
\usepackage{booktabs}
\usepackage{mathtools}
\usepackage{color, colortbl}
\usepackage{float}
\usepackage{multirow}
\usepackage{threeparttable}
\usepackage{fancyvrb}
\usepackage{caption}
\usepackage{soul}
\usepackage{diagbox}
\usepackage{bm}
\usepackage{bbding}
\usepackage{pifont}
\usepackage{wasysym}
\usepackage{microtype}
\definecolor{Gray}{gray}{0.1}
\usepackage[dvipsnames]{xcolor}

\definecolor{cvprblue}{rgb}{0.21,0.49,0.74}
\definecolor{highlightred}{RGB}{255,99,100}
\definecolor{highlightblue}{RGB}{64,140,255}
\definecolor{Highlight2}{RGB}{227,243,255}

\newcommand{\hred}[1]{\textcolor{highlightred}{#1}}
\newcommand{\hblue}[1]{\textcolor{highlightblue}{#1}}

\usepackage[pagebackref,breaklinks,colorlinks,citecolor=cvprblue]{hyperref}

\def\paperID{03240} 
\def\confName{CVPR}
\def\confYear{2024}

\title{Low-Resource Vision Challenges for Foundation Models}

\author{Yunhua Zhang$^{1}$~~~~Hazel Doughty$^{2}$~~~~Cees G. M. Snoek$^{1}$\\[1mm]
\normalsize $^{1}$University of Amsterdam~~~$^{2}$Leiden University \\
}

\begin{document}
\maketitle
\begin{abstract}
\vspace{-0.8em}
Low-resource settings are well-established in natural language processing, where many languages lack sufficient data for deep learning at scale. However, low-resource problems are under-explored in computer vision. In this paper, we address this gap and explore the challenges of low-resource image tasks with vision foundation models. We first collect a benchmark of genuinely low-resource image data, covering historic maps, circuit diagrams, and mechanical drawings. These low-resource settings all share three challenges: data scarcity, fine-grained differences, and the distribution shift from natural images to the specialized domain of interest. While existing foundation models have shown impressive generalizability, we find they cannot transfer well to our low-resource tasks. 
To begin to tackle the challenges of low-resource vision, we introduce one simple baseline per challenge. 
Specifically, we i) enlarge the data space by generative models, ii) adopt the best sub-kernels to encode local regions for fine-grained difference discovery and iii) learn attention for specialized domains. Experiments on our three low-resource tasks demonstrate our proposals already provide a better baseline than transfer learning, data augmentation, and fine-grained methods. This highlights the unique characteristics and challenges of low-resource vision for foundation models that warrant further investigation. 
Project page: \url{https://xiaobai1217.github.io/Low-Resource-Vision/}. 

\end{abstract}    
\vspace{-1em}
\section{Introduction}
\vspace{-0.3em}
\label{sec:intro}

\begin{figure}[t!]
\centering 
\includegraphics[width=1\linewidth]{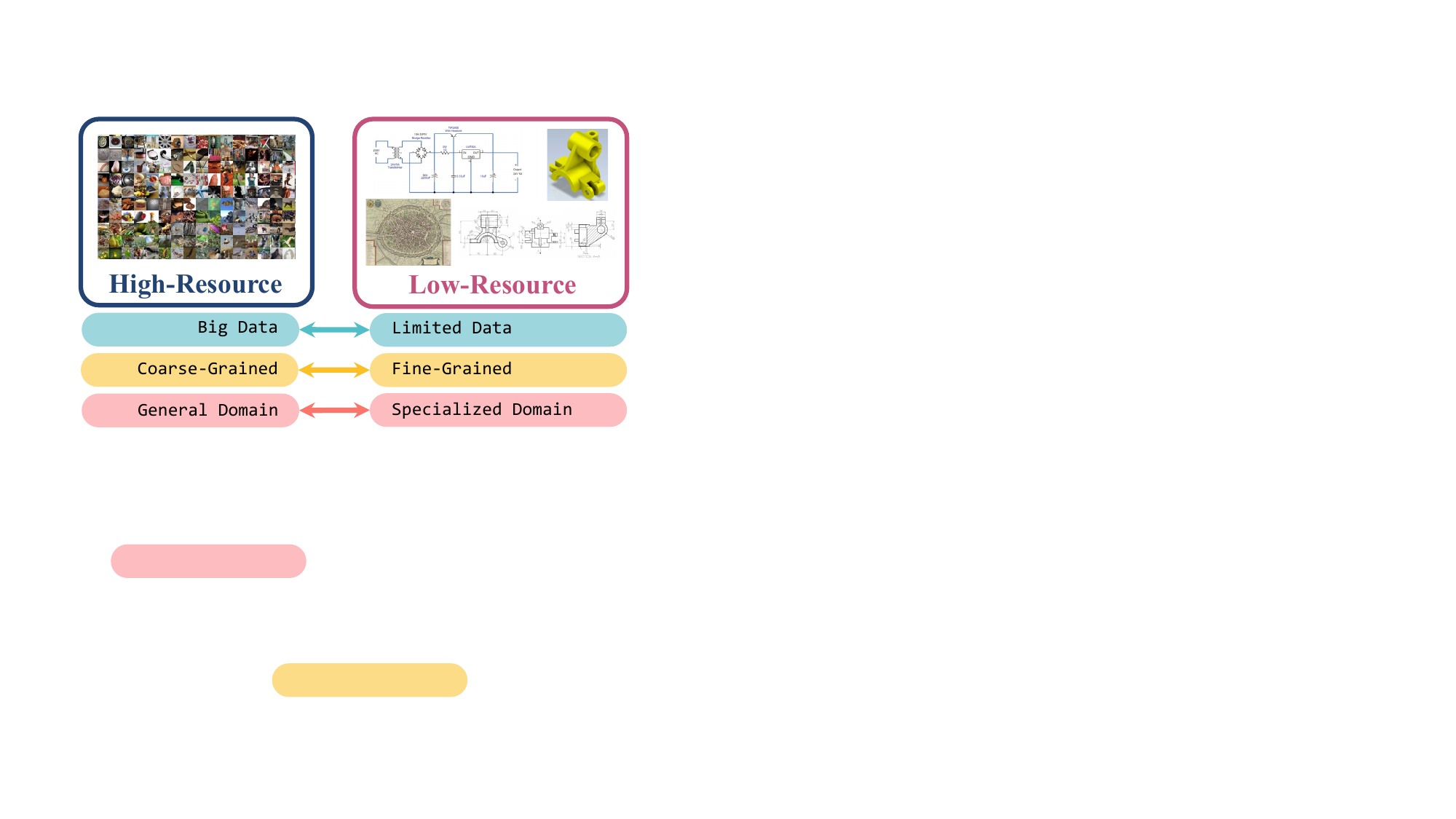}
\vspace{-2em}
\caption{\textbf{High-Resource vs Low-Resource Vision}. High-resource vision focuses on images that can be collected at scale, have coarse-grained classes, and come from a general domain. We instead focus on low-resource vision tasks where data is scarce, has fine-grained differences, and comes from highly specialized domains. 
}
\label{fig:1st_figure}
\vspace{-1.2em}
\end{figure}

Many have studied low-resource natural language processing~\cite{zoph2016transfer,adams-etal-2017-cross,fadaee2017data,pan2017cross,Hedderich2020ASO}, in which the target languages are less common and data is scarce. In computer vision, numerous works have explored effective learning methods for limited labeled data scenarios, \eg, by meta-learning~\cite{hospedales2021meta,Hu_2023_CVPR}, few-shot learning~\cite{vinyals2016matching,Liu_2023_CVPR}, or generative modeling~\cite{fei2006one,Zheng_2023_CVPR}. Albeit successful, they focus on high-resource image domains, where thousands of images from the same domain are available, even though each class may only have a few samples for model learning. Different from existing works handling data scarcity, we investigate low-resource settings for computer vision where data is truly scarce (see Figure~\ref{fig:1st_figure}).

By collecting a benchmark of low-resource vision tasks we are able to study the combination of challenges unique to this area. First, data is severely limited with only a few hundred examples available for model learning. Second, vision tasks that are low-resource also tend to be highly specialized, meaning differences between different images are extremely subtle and fine-grained. Finally, the limited examples and specialized nature of the task means the domain is incredibly different from more common natural images available in bulk. While these challenges are often studied in isolation~\cite{wang2021towards,hong2020trashcan} or even pairs~\cite{wilds,bejnordi2017diagnostic}, the combination of all three is unique to low-resource vision and demands solutions outside of the scope of current models. %

An intuitive way to handle low-resource vision is leveraging the strong image representations from foundation models, which have progressed at a tremendous pace in recent years  \cite{CLIP,BLIP,ALIGN,kirillov2023segment,imagebind}. 
They show promising zero-shot performance on various downstream vision tasks and provide representations with generalization and transfer capabilities. Thus, they are a natural solution to low-resource vision.
However, we find that current foundation models~\cite{CLIP,BLIP,imagebind,li2023blip} struggle to generalize to the specialized domains of low-resource vision tasks. %
We also find that existing transfer learning techniques struggle to adapt with the very limited amount of data available. Thus, we propose several adaptation baselines to begin to tackle the challenges of low-resource vision, with the ambition to inspire future work in this area. %

As our main contribution, we study the challenges of low-resource vision and collect a low-resource image benchmark. Specifically, our benchmark covers circuit diagrams, historic maps, and mechanical drawings. We find the challenges of low-resource vision are a lack of training data, fine-grained differences, and domain shift from natural images to specialized domains. From our analysis, we discover foundation models struggle to recognize and retrieve low-resource images as do existing transfer learning methods. Thus, we introduce three simple baselines to mitigate each difficulty. Specifically, we finetune foundation models using diverse data produced by generative models to cope with data scarcity, we discover fine-grained details by focusing on local patterns via selected sub-kernels, and we learn attention for specialized domains to combat the distribution shift. Experiments demonstrate the challenges of our low-resource benchmark for existing transfer learning, data augmentation, and fine-grained methods as well as the advantages of our baselines, which can be added to different foundation models. We also discuss the remaining challenges for low-resource vision and paths forward for future works.

\vspace{-0.3em}
\section{Low-Resource Image Transfer Evaluation}
\label{sec:LITE}
\vspace{-0.3em}

To study low-resource vision tasks we cannot simply take a subset of existing data and pretend it is low-resource. This will not present the same challenges as are present in true low-resource data. Instead, we collect image data that is severely limited in its online availability. %
Specifically, we present our Low-Resource Image Transfer Evaluation (LITE) benchmark which considers three low-resource vision tasks. We examine the common challenges among these tasks and whether they can be solved by foundation models. 

\vspace{-0.1em}
\subsection{Tasks}
\begin{figure}[t!]
\centering
\includegraphics[width=0.9\linewidth]{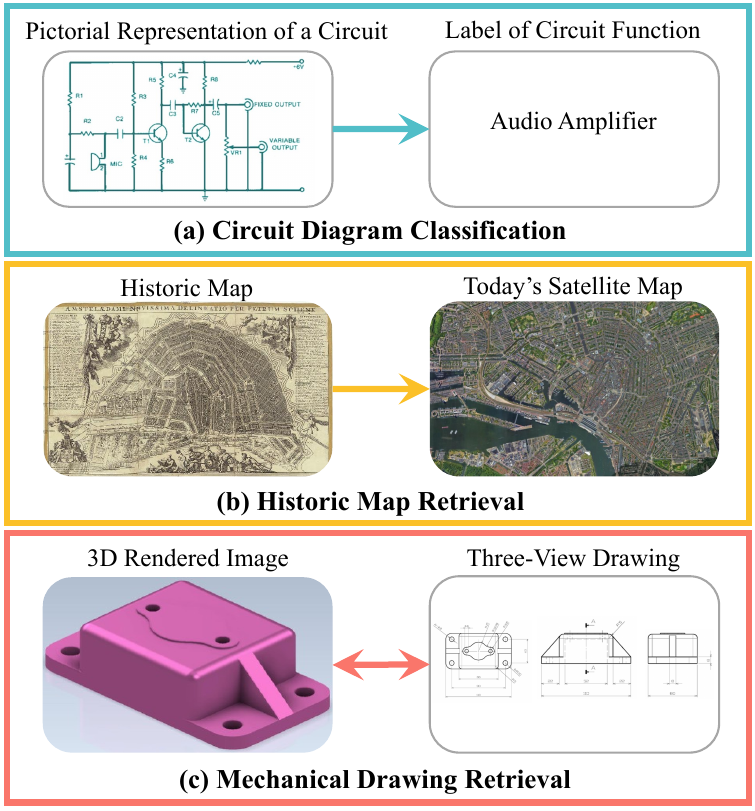}
\vspace{-1em}
\caption{\textbf{Low-Resource Image Transfer Evaluation Benchmark}. 
Our three benchmark tasks are: (a) classifying circuit diagrams with the correct function, (b) retrieving the modern satellite map given an old map of a city, and (c) retrieving the mechanical drawing corresponding to a 3D photo of a component and vice versa.
}
\vspace{-1em}
\label{fig:benchmark}
\end{figure}

\vspace{-0.3em}
Our benchmark has three tasks: (i) circuit diagram classification, (ii) image-to-image retrieval with historic maps, and (iii) image-to-image retrieval with mechanical drawings. Examples from each task are shown in Figure~\ref{fig:benchmark}.%

\noindent \textbf{Task I: Circuit Diagram Classification}. 
The goal is to classify the images of circuit diagrams by their function, \eg, audio amplifier and power supply. We collect circuit images and labels from books~\cite{book} and websites~\cite{Gadgetronicx, circuitdigest,circuitdiy}. %
In total, we have 32 function classes which are equally represented in training. 
The challenge comes from small changes in circuit components dramatically changing the function. Since there are different layouts for the same function, it is also easy for models to overfit to specific layouts. We measure performance with Top-1 and Top-5 accuracy.

\noindent \textbf{Task II: Historic Map Retrieval}. The task is to retrieve the corresponding modern-day satellite image for each image of a historic city map. Data is acquired from Old Maps Online~\cite{oldmaps} and cropping the corresponding contemporary satellite image from Google Maps~\cite{googlemap}. This task is challenging as many city layouts have changed considerably over time and the contours of walls and buildings in the historic map may no longer exist in the satellite image. Moreover, historic maps originating from different regions and eras have vastly different cartographic styles. Performance is measured using Recall@1, Recall@5, and mean rank. 

\noindent \textbf{Task III: Mechanical Drawing Retrieval}. 
The goal is to retrieve the mechanical drawing matching the image of a 3D-rendered component and vice versa. We collect mechanical drawings and rendered images from TraceParts~\cite{traceparts} and GrabCAD~\cite{grabcad}.
The difficulty comes from the large visual difference between image sets. Moreover, the mechanical drawings and rendered images use different viewpoints. We evaluate this task with Recall@1, Recall@5, and mean rank, each averaged across both retrieval directions. 

We summarize our benchmark statistics in Table~\ref{tab:datasets}. Note that we have collected as much data as we can find freely available online for each task, yet, the amount of data is still incredibly small showing how low-resource these tasks are. 

\begin{table}[t!]
\setlength{\tabcolsep}{2pt}
\centering
\resizebox{1\linewidth}{!}{
\begin{tabular}{llrrr}
\toprule

\textbf{Task} & \textbf{Formulation} & \textbf{Train} & \textbf{Val} & \textbf{Test} \\
\midrule
\textbf{Circuit Diagram Classification} & Image Classification & 154 & 100 & 1,078 \\
\textbf{Historic Map Retrieval} & Image-to-Image Retrieval & 102  & 140  & 409 \\
\textbf{Mechanical Drawing Retrieval} & Image-to-Image Retrieval & 300  & 100  & 754  \\
\bottomrule
\end{tabular}
}
\vspace{-0.8em}
\caption{\textbf{LITE Benchmark Statistics.} 
We show the task formulation and number of images (or image pairs) per split for each of our three tasks. The benchmark is available on our project website. %
}
\vspace{-1em}
\label{tab:datasets}
\end{table}

\vspace{-0.1em}
\subsection{Low-Resource Vision Challenges}
\label{sec:challenges}
\vspace{-0.3em}

While the three low-resource tasks forming our LITE benchmark are very diverse, we identify three common challenges. %

\noindent \textbf{Challenge I: Data Scarcity. }
The data available for training models for low-resource scenarios is extremely limited. This is demonstrated through the small amount of data we were able to find online for each low-resource task (see Table~\ref{tab:datasets}).

\noindent \textbf{Challenge II: Fine-Grained. }
Data that is low-resource is also highly specialized, meaning differences between images are incredibly subtle and attention to fine-grained details is necessary to solve the task. For example, the component symbols are key to a circuit's purpose, not its layout. Similarly, in mechanical drawings, the components may only vary in the number of holes. 

\noindent \textbf{Challenge III: Specialized Domain. }
Not only is the available data severely limited, but it has a significantly different appearance to the natural images commonly used in vision tasks. This means it is difficult to bootstrap the training data for low-resource tasks with 
existing datasets. Moreover, models that are successful on natural images cannot be easily applied to the specialized domains of low-resource images.  %

Each of these challenges has been studied in isolation in vision, for instance with few-shot learning~\cite{wang2020generalizing,song2023comprehensive}, fine-grained classification~\cite{wang2021max,zhu2022dual,he2022transfg} and domain generalization~\cite{wang2022generalizing}. However, their combination is unique to low-resource vision tasks. This means existing solutions to individual challenges cannot be easily applied to low-resource vision.
Considering these challenges and their combination, we identify foundation models as the existing solution with the most potential to tackle low-resource vision, due to the impressive generalizability foundation models have shown. 
In the following section, we propose one way to better adapt foundation models for each low-resource vision challenge.

\vspace{-0.3em}
\section{Baselines for the Low-Resource Challenges}
\label{sec:challenges}
\vspace{-0.3em}
Our goal is to adapt foundation models, pre-trained on large-scale datasets, to low-resource tasks. A foundation model $\mathcal{F}$ can be adapted with a small set of transfer learning parameters $\bm{\theta}$ as in LoRA~\cite{hu2021lora} or AdaptFormer~\cite{adaptformer}. To better handle adaptation in low-resource vision, we introduce one baseline for each challenge highlighted in Section~\ref{sec:LITE}. 
First, to cope with the lack of data, we propose to augment training samples via generative models 
(Section~\ref{sec:diverse}). 
Second, to focus on the fine-grained details, we reduce the token patch size with selective tokenization %
(Section~\ref{sec:selective}).
Third, we introduce attention for specialized domains for better model adaption %
(Section~\ref{sec:adaptive}). 
During finetuning on a low-resource task, we fix the foundation model $\mathcal{F}$ and train the parameters for transfer learning, our tokenization, and our attention. 


\vspace{-0.1em}
\subsection{Baseline I: Generated Data for Data Scarcity} \label{sec:diverse}
\vspace{-0.3em}
\textbf{Objective.} Since a major challenge of low-resource vision is data scarcity, models are prone to overfitting the training data. We address this challenge by creating more training data for a low-resource vision task through generative models.

\noindent\textbf{Novelty.} Prior works have used generative models to produce realistic images to augment the training data~\cite{trabucco2023effective,he2022synthetic}. However, these works focus exclusively on data where the label of the augmented image is known. With this approach, it is challenging to achieve good data diversity with the highly limited number of images available in low-resource vision tasks. Therefore, besides label-preserving images, we use images where the original label is broken and unknown.

\begin{figure}[t!]
\centering
\includegraphics[width=1\linewidth]{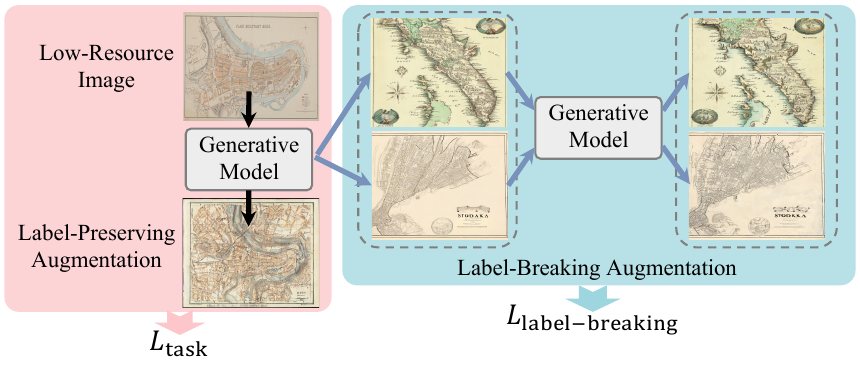}
\vspace{-2.2em}
\caption{
\textbf{Generated Data for Data Scarcity}. We augment images with generative models, obtaining images close to the input image where the label is preserved as well as more diverse images which break the label. We use label-preserving images in the task loss and augment the label-breaking images for use in a contrastive loss.
}
\vspace{-1em}
\label{fig:data_aug}
\end{figure}

\noindent\textbf{Method. }
Our proposed baseline is shown in Figure~\ref{fig:data_aug}. 
In Stable Diffusion~\cite{stablediffusion}, the forward process gradually adds Gaussian noise to an image with a variance schedule $\beta_1, ..., \beta_T$ ($T{=}50$). 
To obtain new images, we sample noisy images at different timestep $t$ and start the reverse process. 
For label-preserving augmentations we want a small $t$ so the generated image is close to the original and thus adopt $\gamma {=} 0.3$ and $t{=}\gamma \cdot T$. For label-breaking augmentations we use $\tau {=} 0.6$ and $t{=}\tau \cdot T$. 
Then, we can obtain various augmented images $[\mathcal{I}_1, \mathcal{I}_2, \cdots, \mathcal{I}_m]$. 
Since the ground truth for the label-breaking augmentation is unknown we instead utilize such data with a contrastive learning objective~\cite{he2020momentum,chen2020simple}. 
To construct the positive pairs, we generate a second augmented image $\mathcal{I}_{j}'$ for each label-breaking augmentation $\mathcal{I}_j$ with sampling timestep $t{=}\gamma \cdot T$. 
The contrastive loss encourages the feature of $\mathcal{I}_{j}'$ to be close to that of the label-breaking image $\mathcal{I}_j$, but far away from other label-breaking augmentations.  
We pass the label-breaking augmentation pairs through the foundation model to obtain their features $x_j{=}\mathcal{F}(\mathcal{I}_j)$ and $x_j'{=}\mathcal{F}(\mathcal{I}_j')$, and our objective becomes:
\vspace{-0.4em}
\begin{equation}
\vspace{-0.4em}
    L_{\text{label-breaking}} = -\frac{1}{N} \sum_{j}^N \text{log} \frac{\text{exp}({\mathbf{x}_{j}'}^{\text{T}} \mathbf{x}_j / \sigma)  }{\sum_{i=1}^N \text{exp}({\mathbf{x}_{j}'}^{\text{T}} \mathbf{x}_i / \sigma)},
\end{equation}
where $N$ is the number of label-breaking image pairs and $\sigma$ is the temperature for logit scaling. Combining this with the original task loss $L_{task}$ our overall learning objective is: 
\vspace{-0.4em}
\begin{equation}
\vspace{-0.4em}
\label{eq:loss}
    L = L_{\text{task}} + \lambda L_{\text{label-breaking}},
\end{equation}
where $\lambda$ is a hyperparameter to balance the loss terms. 
For $L_{\text{task}}$ we use the original images as well as the label-preserving augmentations in a softmax cross-entropy for classification and a contrastive loss for retrieval. 
During each training iteration, we randomly sample $B$ images from the union set of original and label-preserving augmentations for the task loss in addition to $B$ image pairs for the label-breaking loss. Following~\cite{he2020momentum} we use a memory bank for the contrastive loss $L_{\text{label-breaking}}$ so that there are $N$ pairs ($N{>}B$) in total. 
As both the label-preserving and the label-breaking augmentations enlarge the data space for model learning, the challenge of limited data can be alleviated considerably. 

\noindent\textbf{Details.} %
We use $\gamma{=}0.3$ as the noise threshold for producing label-preserving augmentations and $\tau{=}0.6$ for label-breaking augmentations. 
We use a batch size of $B{=}8$ and generate $m{=}10$ augmented images per sample. 
The contrastive learning uses $N{=}100$ as the memory bank size and loss terms are balanced with $\lambda{=}0.1$ (Eq.~\ref{eq:loss}). 
We optimize the losses using Adam~\cite{kingma2014adam} with a learning rating of $10^{-3}$. The model is trained on one NVIDIA A6000 for $90$ epochs.

\vspace{-0.1em}
\subsection{Baseline II: Tokenization for Fine-Grained} \label{sec:selective}
\vspace{-0.3em}
\textbf{Objective. }The second major challenge in low-resource vision, is the subtle, fine-grained details that distinguish different images. To address this challenge, we simply reduce the image patch size for tokenization so that the model can attend to the finer details of a low-resource input image.

\begin{figure}[t!]
\centering 
\includegraphics[width=1\linewidth]{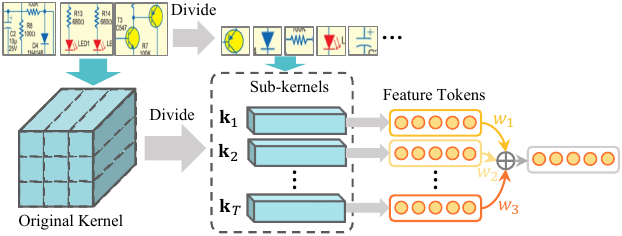}
\vspace{-2em}
\caption{\textbf{Tokenization for Fine-Grained}.
We divide the original linear projection of a pre-trained foundation model into sub-kernels. These sub-kernels can be applied to smaller areas of the image patch to attend to fine-grained details. We learn a weighting to combine the resulting features into patch-level features. }
\label{fig:tokenization}
\vspace{-1.2em}
\end{figure}

\noindent\textbf{Novelty. }
As we have limited data, we cannot train a new tokenization layer from scratch to reduce patch size. 
As shown in Figure~\ref{fig:tokenization}, we instead divide the original linear projection kernel into sub-kernels which can be applied to smaller image patches. We then create patch-level features with a learned weighting. This allows attention to be paid to small local regions, crucial for fine-grained recognition~\cite{ding2019selective,zheng2017learning,du2020fine}, while only adding a handful of parameters. 

\noindent\textbf{Method. }
Vision foundation models~\cite{CLIP,BLIP,kirillov2023segment,imagebind} divide the input image into large patches, \eg, $16 {\times} 16$ or $14 {\times} 14$, so that the number of resulting tokens is small allowing training with large batch sizes. 
These image patches are linearly projected into features. The mechanism for this linear projection can be viewed as a convolution kernel $\mathbf{K} \in \mathbb{R}^{q \times q {\times} 3 {\times} d_{\text{model}}}$ where $q {\times} q {\times} 3$ is also the dimensionality of an input image patch. %
We divide this kernel $\mathbf{K}$ into a series of sub-kernels $\{\mathbf{k}_1, \cdots, \mathbf{k}_T\, | \mathbf{k}_t \in \mathbb{R}^{u {\times} u {\times} 3 {\times} d_{\text{model}}} \}$, where $u{<}q$.
We use each sub-kernel to encode an input image patch and obtain a series of features, one per sub-kernel, $\{\mathbf{b}_1, \cdots, \mathbf{b}_T | \mathbf{b}_t \in \mathbb{R}^{p {\times} p {\times} d_{\text{model}}} \}$, where $p$ is the size of feature maps. Unlike the original linear projection, the sub-kernels can find smaller, fine-grained patterns in the input image patch, achieving a similar effect to a reduced patch size. We learn a weighting $\mathbf{w}{=}[w_1, \cdots, w_T]$ to combine sub-kernel features into a patch-level feature $\mathbf{b}$:
\vspace{-0.4em}
\begin{equation}
\vspace{-0.4em}
    \mathbf{b} = \sum_t \frac{e^{w_t}}{\sum_t e^{w_t}} \mathbf{b}_t.
\end{equation}
To obtain output features with the same dimension as the original projection, we apply max pooling $\mathbf{b}' {=} \text{MaxPool}(\mathbf{b})$ and flatten $\mathbf{b'}$. We can then use the existing positional encodings and class tokens and ensure our input is suitable for the frozen foundation model. As only the weighting $\mathbf{w} \in \mathbb{R}^T$ is learned, our tokenization is suitable for low-resource data and effectively encourages focus on fine-grained details. 

\noindent\textbf{Details.} The initial kernel has size $q{=}14$. 
We set $u{=}7$, giving $T{=}49$ sub-kernels. 
The resulting feature maps of size $p{=}32$ are max pooled with a kernel size and stride of 2.

\vspace{-0.1em}
\subsection{Baseline III: Attention for Specialized Domains} \label{sec:adaptive}
\vspace{-0.3em}
\textbf{Objective.} 
The third challenge considers the adaption of foundation model features to the specialized low-resource domains. 
The transformer attention of foundation models struggles to distinguish the important regions in our specialized low-resource domains, we thus propose an alternative. 

\begin{figure}[t!]
\centering 
\includegraphics[width=0.8\linewidth]{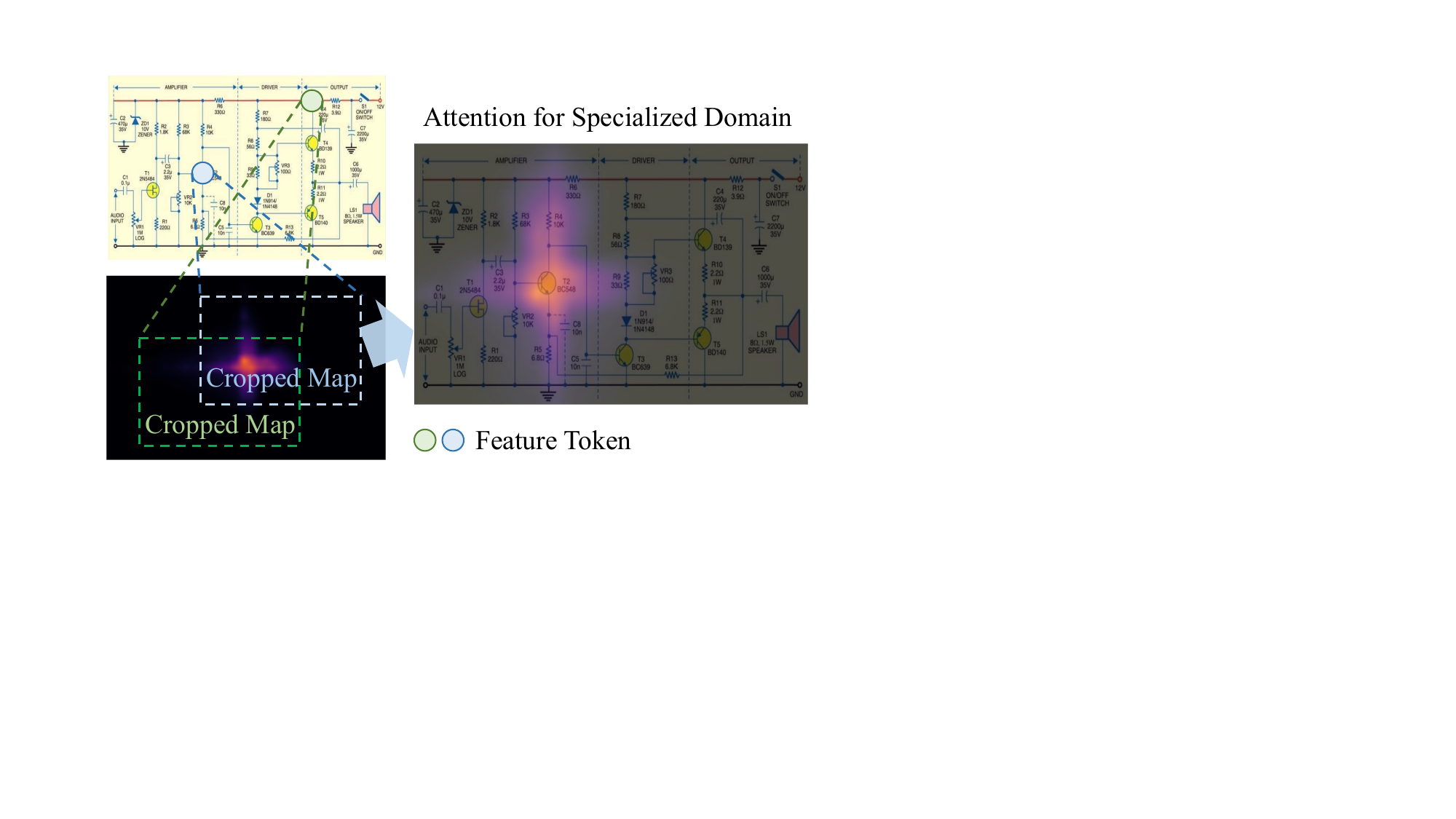}
\vspace{-0.8em}
\caption{\textbf{Attention for Specialized Domains}. We learn a set of global attention maps with common attention patterns particular to the specialized domain such as vertical and horizontal directions for circuit diagrams. For each token, we crop the corresponding region from the global attention map according to the location. 
}
\label{fig:att_crop}
\vspace{-1em}
\end{figure}

\noindent \textbf{Novelty.}
We observe that the type of attention required is specific to each domain, but can be shared across different images and different patches within an image. For example, the vertical and the horizontal surroundings of a patch are important in circuit diagrams, while for historic maps, local neighbors are essential. To reduce the number of parameters, we share attention maps across samples and feature tokens by learning global attention maps. As shown in Figure~\ref{fig:att_crop}, for each feature token, we simply crop the corresponding attention maps from the global maps. %

\noindent \textbf{Method.}
Specifically, we learn $C$ attention maps $\bm{\mathcal{M}} \in \mathbb{R}^{C {\times} 2h {\times} 2h}$, where $h$ is the height and width of the feature maps before being flattened for input into the following transformer blocks. 
Each attention map will correspond to a different attention pattern. 
To obtain the correct size of $h {\times} h$ for a token's attention map we crop a sub-map from each global attention map. For a token corresponding to location $(i,j)$, the top-left corner in the global attention map is $(h{-}i, h {-} j)$. %
As a result, we obtain one $h {\times} h$ sub-map for each of the $h^2$ tokens and form $\bm{\mathcal{M}}_c' \in \mathbb{R}^{h^2 {\times} h {\times} h}$, for each of the $C$ global attention maps. 
We flatten $\bm{\mathcal{M}}_c'$ into $\bm{\mathcal{M}}_c \in \mathbb{R}^{h^2 \times h^2}$, and apply softmax to the last dimension. 
The resulting attention is multiplied with the values $\mathbf{V}$ used in the original multi-head self-attention. We weigh the resulting features with a learned vector $\mathbf{r}{=}[r_1, \cdots, r_C]$ as follows:
\vspace{-0.4em}
\begin{equation}
\vspace{-0.4em}
    \mathbf{\hat{f}}_l = \sum_{c=0}^{C} \frac{e^{r_c}}{\sum_c e^{r_c}} \text{MLP}(\text{softmax}(\bm{\mathcal{M}}_c) \mathbf{V}),
\end{equation}
where the multi-layer perceptron (MLP) is the same as used in the transformer layer's multi-head attention.  
We combine the output from our attention for specialized domains $\mathbf{\hat{f}}_l$ with the output from multi-head attention $\mathbf{\bar{f}}_l$ as: 
\vspace{-0.4em}
\begin{equation}
\vspace{-0.4em}
    \mathbf{\bar{f}}_l' = \mathbf{\bar{f}}_l + \alpha \mathbf{\hat{f}}_l,
\end{equation}
where $\alpha$ is learned to balance the two attentions. 
As only $\bm{\mathcal{M}}$, $\alpha$, and $\mathbf{r}$ are learned, training our attention for specialized domains allows adaptation without overfitting. 

\noindent\textbf{Details.} We learn $C{=}10$ maps for the middle ($16$th) transformer block, leaving other blocks unchanged.

\vspace{-0.3em}
\section{Related Work}
\vspace{-0.3em}
%

\begin{table*}[t!]
\centering
\resizebox{0.8\linewidth}{!}{
\begin{tabular}{lcccccccc}
\toprule
 & \multicolumn{2}{c}{\textbf{Circuit Diagram Classification}} & \multicolumn{3}{c}{\textbf{Historic Map Retrieval}} & \multicolumn{3}{c}{\textbf{Mechanical Drawing Retrieval}} \\
 \cmidrule(lr){2-3} \cmidrule(lr){4-6} \cmidrule(lr){7-9} 
 & Top-1 (\%) $\uparrow$ & Top-5 (\%) $\uparrow$ & R@1 $\uparrow$ & R@5 $\uparrow$ & MnR $\downarrow$ & R@1 $\uparrow$ & R@5 $\uparrow$ & MnR $\downarrow$ \\
\midrule
\textbf{CLIP}~\cite{CLIP} & 7.7 & \textbf{\hblue{28.5}} & \textbf{\hred{31.3}} & \textbf{\hblue{60.4}} & \textbf{\hblue{12.1}} & 3.6 & 10.5 & 210.2 \\
\textbf{BLIP}~\cite{BLIP} & \textbf{\hblue{8.7}} & 28.2 & 2.2 & 12.5  & 52.1 & 2.5 & 7.8 & 209.4 \\
\textbf{SAM}~\cite{kirillov2023segment} & - & - & 0.7 & 3.2 & 97.0 & 0.1 & 0.8 & 369.2 \\
\textbf{AIM}~\cite{el2024scalable} & - & - & 12.0 & 33.0 & 37.9 & \textbf{\hblue{14.9}} & \textbf{\hblue{31.2}} & 72.2 \\
\textbf{DINOv2}~\cite{oquab2023dinov2} & - & - & 1.5 & 7.1 & 83.4 & \textbf{\hred{15.9}} & \textbf{\hred{32.2}} &  \textbf{\hred{83.0}} \\
\textbf{ImageBind}~\cite{imagebind} & \textbf{\hred{19.3}} & \textbf{\hred{45.1}} & \textbf{\hblue{28.1}} & \textbf{\hred{62.1}} &  \textbf{\hred{10.1}} & 13.2 & 26.3 &  \textbf{\hblue{83.1}} \\
\bottomrule
\end{tabular}
}
\vspace{-1em}
\caption{
\textbf{Difficulties for Vision Foundation Models}. 
We present zero-shot transfer performance. 
We mark the best in \textbf{\hred{red}} and the second in \textbf{\hblue{blue}}. While ImageBind~\cite{imagebind} has generally better zero-shot transfer ability on low-resource vision tasks, the tasks are far from solved. 
}
\vspace{-1.2em}
\label{tab:foundation_models}
\end{table*}

\noindent\textbf{High-Resource Vision}. 
The large majority of computer vision research focuses on high-resource settings, where data is plentiful. Various benchmarks of high-resource images have been proposed, \cite{russakovsky2015imagenet,lin2014microsoft,cordts2016cityscapes,geiger2012we,netzer2011reading,liu2015deep,krishna2017visual}, unlocking the ability to train larger and larger models. 
Their images are crawled from the internet~\cite{russakovsky2015imagenet,lin2014microsoft,liu2015deep,krishna2017visual}, or captured by the authors~\cite{geiger2012we,cordts2016cityscapes,netzer2011reading}. 
The labels can be either coarse-grained~\cite{russakovsky2015imagenet,lin2014microsoft,geiger2012we,cordts2016cityscapes,ypsilantis2023towards,bastani2023satlaspretrain} or fine-grained~\cite{liu2015deep,krishna2017visual,netzer2011reading,yang2023emoset,khan2023fishnet}. 
High-resource vision tends to focus on natural images which are plentiful online. However, some benchmarks also collect images from other domains, \eg, X-ray~\cite{wang2021towards}, underwater~\cite{hong2020trashcan}, medical~\cite{irvin2019chexpert} and satellite~\cite{helber2019eurosat}. These are less high-resource than natural images. However, they still contain thousands of samples. Different from previous high-resource image datasets, we focus on low-resource settings, where images are severely limited with only a few hundred samples available for training.

\noindent \textbf{Vision Foundation Models. } 
Vision foundation models are pre-trained by high-resource web-crawled images with weak supervision or human annotations, and present impressive generalizability on various downstream tasks. 
While CLIP~\cite{CLIP}, BLIP~\cite{BLIP}, and ALIGN~\cite{jia2021scaling} learn from image-text pairs only, ImageBind~\cite{imagebind} uses image-paired data of multiple modalities. 
Recent works SAM~\cite{wang2023detecting}, DINOv2~\cite{oquab2023dinov2}, UniDetector~\cite{kirillov2023segment} and AIM~\cite{el2024scalable} instead propose foundation models for visual-only tasks such as object detection, segmentation and depth estimation. %
However, the impressive generalization ability has been focused on natural images, likely similar to many in the large-scale training set~\cite{xu2023demystifying}. Simultaneously, there are many low-resource problems from specialized domains lacking a large amount of online data, such as technical images. In this paper, we create a benchmark of low-resource vision problems and demonstrate that foundation models cannot generalize to such data. 

It is also possible to adapt the strong image representations of foundational models to new tasks. %
This can be done by fintuning~\cite{yosinski2014transferable} or by training additional projection layers~\cite{clipadapter}. 
Several works~\cite{hu2021lora,adaptformer,he2023sensitivity,liu2022few,toast} instead add new trainable parameters into the layers of a frozen pre-trained model. 
Although these works achieve impressive performance, they are not suited for low-resource vision where training data is severely limited and from fine-grained, specialized domains that are highly dissimilar to the pre-training data. We study such tasks and their challenges and propose baselines for better adaptation to low-resource tasks.

\noindent\textbf{Low-Shot Vision}. %
A huge number of works have studied scenarios with limited training data. One typical setting is few-shot learning~\cite{vinyals2016matching, omniglot, bertinetto2018meta}, which aims to generalize to previously unseen classes with only a few training samples. 
Some works study in-context learning~\cite{zhang2023makes,bar2022visual,wang2023context,sun2023imagebrush,sun2023exploring}, which allows inference on unseen tasks by conditioning on related examples without updating the model parameters. 
Other works reduce this one step further and study zero-shot scenarios~\cite{akata2015label,xian2017zero,xian2018feature,zhang2017learning,changpinyo2016synthesized,kodirov2017semantic}, where no data of the relevant classes are seen in training, although prior knowledge or data from other classes could be used. 
All these works make a significant step towards reducing the amount of data needed for model learning. However, none of these tasks study the combination of scarce data, fine-grained differences and highly specialized domains present in low-resource vision.

\vspace{-0.3em}
\section{Results and Discussion}
\vspace{-0.3em}

\vspace{-0.1em}
\subsection{Difficulties for Vision Foundation Models}
\vspace{-0.3em}

To understand how well current vision foundation models address low-resource vision tasks, we first examine their zero-shot performance on our LITE benchmark. We consider six vision foundation models: CLIP~\cite{CLIP}, BLIP~\cite{BLIP}, SAM~\cite{kirillov2023segment}, AIM~\cite{el2024scalable}, DINOv2~\cite{oquab2023dinov2} and ImageBind~\cite{imagebind}. %

\noindent \textbf{Setup.}
To obtain zero-shot results for circuit diagram classification, we follow \cite{CLIP} and customize the label text to make it better suited to the models. Specifically, we use the prompt template ``A circuit diagram of \{label\}.'', where the label is a category label, \eg, power supply or motor driver. 
We cannot obtain zero-shot results for SAM, AIM, and DINOv2 in this way, so we omit these models for circuit diagram classification. 
For all tasks, we calculate the similarities among the feature embeddings between the input image and the ground-truth image or text to find the closest neighbors.

\noindent \textbf{Results.} %
From  Table~\ref{tab:foundation_models} we observe none of these foundation models perform well on low-resource tasks. 
Although ImageBind is better suited due to its larger pre-training set and image-focused embedding, there is still much room for improvement. 
Despite current foundation models' impressive generalizability on other benchmarks, they cannot yet solve the combined challenges of data scarcity, fine-grained details, and highly specialized domains. 
Unlike other zero-shot and few-shot tasks where foundation models have shown good generalization, low-resource data is truly scarce online, meaning it is unlikely to be in the training data of foundation models. The specialized domain means it is also highly dissimilar to natural images which form a large part of foundation model training data~\cite{xu2023demystifying}. Due to the models' unfamiliarity with low-resource data, they struggle to attend to fine-grained, task-relevant details. Therefore, vision foundation models need adaptation for low-resource tasks. 

\begin{table}[t!]
\centering
\resizebox{1\linewidth}{!}{
\begin{tabular}{lccc}
\toprule
 & \multicolumn{2}{c}{\textbf{Circuit Classification}} \\
 \cmidrule(lr){2-3}
 & Top-1 (\%) $\uparrow$ & Top-5 (\%) $\uparrow$ \\ %
\midrule
Zero-Shot Transfer & 19.3 & 45.1 \\
\rowcolor{gray!20}
\multicolumn{3}{l}{\textbf{Simple Transformations}}  \\
Random Crop and Flip & 19.8 & 45.3\\
Mixup~\cite{zhang2017mixup} & 20.8 & 46.0 \\
CutMix~\cite{yun2019cutmix} & 20.0 & 45.5 \\
Random Erasing~\cite{zhong2020random} & 20.8 & 46.2 \\
\rowcolor{gray!20}
\multicolumn{3}{l}{\textbf{Generative Models}}  \\
DA-Fusion~\cite{trabucco2023effective} & 19.6 & 45.1 \\ %
SyntheticData~\cite{he2022synthetic} & 20.8 & 46.0 \\ %
\rowcolor{gray!20}
\multicolumn{3}{l}{\textbf{\textit{Our Baselines}}}  \\
Generated Data for Data Scarcity & \textbf{\hblue{21.3}} & \textbf{\hblue{46.9}} \\
Combination of Baselines & \textbf{\hred{24.1}} & \textbf{\hred{49.3}} \\
\bottomrule
\end{tabular}
}
\vspace{-1em}
\caption{\textbf{Challenge I: Data Scarcity}. We mark the best in \textbf{\hred{red}} and the second in \textbf{\hblue{blue}}. Simple transformations do little to improve the diversity of training data. We obtain the best data diversity and thus the best baseline performance with our baselines which leverage both similar and dissimilar images produced by generative models. 
}
\vspace{-1em}
\label{tab:augmentation}
\end{table}

\vspace{-0.1em}
\subsection{Challenge Results} 
\label{sec:challenge_results}
\vspace{-0.3em}

\noindent \textbf{Setup.} As ImageBind obtains the best zero-shot performance on our low-resource benchmark, we use it in this section. For all three challenges defined in Section~\ref{sec:LITE} we add our proposed baselines or existing methods alongside AdaptFormer~\cite{adaptformer}, keeping the foundation model frozen. 
Our baselines are independent of each other, focusing on different areas of the foundation model: input, tokenization, and attention. Thus they can be easily combined. We test this combination as well as the individual baselines.

\begin{table}[t!]
\centering
\resizebox{0.95\linewidth}{!}{
\begin{tabular}{lccc}
\toprule
 & \multicolumn{2}{c}{\textbf{Circuit Classification}} \\
 \cmidrule(lr){2-3} 
 & Top-1 (\%) $\uparrow$ & Top-5 (\%) $\uparrow$ \\%
\midrule
Zero-Shot Transfer & 19.3 & 45.1\\
\rowcolor{gray!20}
\multicolumn{3}{l}{\textbf{Fine-Grained}}  \\
Adaptive-FGSBIR~\cite{bhunia2022adaptive} & 16.7 & 43.2 \\ 
PLEor~\cite{wang2023open} & 17.1 & 44.1 \\ 
PDiscoNet~\cite{van2023pdisconet} & 16.2 & 43.5 \\ 
\rowcolor{gray!20}
\multicolumn{3}{l}{\textbf{\textit{Our Baselines}}}  \\
Tokenization for Fine-Grained & \textbf{\hblue{20.9}} & \textbf{\hblue{45.5}} \\
Combination of Baselines & \textbf{\hred{24.1}} & \textbf{\hred{49.3}} \\
\bottomrule
\end{tabular}
}
\vspace{-1em}
\caption{\textbf{Challenge II: Fine-Grained}. The \textbf{\hred{best}} and \textbf{\hblue{second}} are highlighted. Fine-grained recognition methods need thousands of images for model learning, making them unsuited to low-resource tasks. Our tokenization baseline better utilizes the limited training data. However, there is much potential for further improvements. 
}
\vspace{-1em}
\label{tab:fine_grained}
\end{table}

\begin{table*}[t!]
\centering
\resizebox{0.83\linewidth}{!}{
\begin{tabular}{lcccccccc}
\toprule
 & \multicolumn{2}{c}{\textbf{Circuit Diagram Classification}} & \multicolumn{3}{c}{\textbf{Historic Map Retrieval}} & \multicolumn{3}{c}{\textbf{Mechanical Drawing Retrieval}} \\
 \cmidrule(lr){2-3} \cmidrule(lr){4-6} \cmidrule(lr){7-9} 
 & Top-1 (\%) $\uparrow$ & Top-5 (\%) $\uparrow$ & R@1 $\uparrow$ & R@5 $\uparrow$ & MnR $\downarrow$ & R@1 $\uparrow$ & R@5 $\uparrow$ & MnR $\downarrow$ \\
\midrule
\rowcolor{gray!20}
\multicolumn{9}{l}{\textbf{CLIP}~\cite{CLIP}}  \\
Zero-Shot Transfer & 7.7 & 28.5 & 31.3 & 60.4 & 12.1 & 3.6 & 10.5 & 210.2 \\
AdaptFormer & 14.0 & 40.1 & 33.3 & 58.4  & 12.3 & 37.3 & 61.5  & 20.3 \\
Our Baselines & \textbf{19.4} & \textbf{45.3} & \textbf{37.1} & \textbf{66.5}  & \textbf{10.9} & \textbf{42.0} & \textbf{66.9}  & \textbf{17.5} \\
\rowcolor{gray!20}
\multicolumn{9}{l}{\textbf{BLIP}~\cite{BLIP}}  \\
Zero-Shot Transfer & 8.7 & 28.2 & 2.2 & 12.5  & 52.1 & 2.5 & 7.8 & 209.4 \\
AdaptFormer & 9.3 & 30.0 & 14.5 &  17.6  & 50.2 & 12.3 & 17.5 &  163.0 \\
Our Baselines & \textbf{14.1} & \textbf{36.7} & \textbf{19.3} & \textbf{20.5}  & \textbf{47.2} & \textbf{17.1} & \textbf{22.6} & \textbf{143.2} \\
\rowcolor{gray!20}
\multicolumn{9}{l}{\textbf{SAM}~\cite{kirillov2023segment}}  \\
Zero-Shot Transfer & - & - & 0.7 & 3.2 & 97.0 & 0.1 & 0.8 & 369.2 \\
AdaptFormer & 18.0 & 44.6 & 8.6 & 12.8  & 59.3 & 7.4 & 10.2 & 221.3 \\
Our Baselines & \textbf{22.4} & \textbf{47.8} & \textbf{13.4} & \textbf{16.3}  & \textbf{55.4} & \textbf{11.2} & \textbf{16.0} & \textbf{178.7} \\
\rowcolor{gray!20}
\multicolumn{9}{l}{\textbf{AIM}~\cite{el2024scalable}}  \\
Zero-Shot Transfer & - & - & 12.0 & 33.0 & 37.9 & 14.9 & 31.2 & 72.2 \\
AdaptFormer & 16.3 & 41.8 & 16.4 & 37.8 & 32.5 & 55.2 & 78.3 & 12.7 \\
Our Baselines & \textbf{20.1} & \textbf{45.7} & \textbf{20.3} & \textbf{41.2} & \textbf{28.8} & \textbf{59.4} & \textbf{82.9} & \textbf{10.0} \\
\rowcolor{gray!20}
\multicolumn{9}{l}{\textbf{DINOv2}~\cite{oquab2023dinov2}}  \\
Zero-Shot Transfer & - & - & 1.5 & 7.1 & 83.4 & 15.9 & 32.2 &  83.0 \\
AdaptFormer & 15.8 & 40.3 & 13.6 & 16.9 & 58.5 & 56.0 & 79.1 & 16.5 \\
Our Baselines & \textbf{20.3} & \textbf{45.8} & \textbf{18.2} & \textbf{21.9} & \textbf{54.1} & \textbf{60.7} & \textbf{83.1} & \textbf{12.4} \\
\rowcolor{gray!20}
\multicolumn{9}{l}{\textbf{ImageBind}~\cite{imagebind}}  \\
Zero-Shot Transfer & 19.3 & 45.1 & 28.1 & 62.1 & 10.1 & 13.2 & 26.3 & 83.1  \\
AdaptFormer & 19.8 & 45.5 & 30.3 & 62.6  & 13.4 & 54.3 & 76.6 & 13.8 \\
Our Baselines & \textbf{24.1} & \textbf{49.3} & \textbf{36.4} & \textbf{68.0} & \textbf{9.8} & \textbf{60.0} & \textbf{82.5} & \textbf{10.2} \\
\bottomrule
\end{tabular}
}
\vspace{-1em}
\caption{\textbf{Combination with Different Foundation Models}. Our approach can easily be applied to different foundational models, improving their adaptation to low-resource tasks. However, the tasks are far from solved highlighting the need for further study of low-resource vision. 
}
\vspace{-1.5em}
\label{tab:foundation_exp}
\end{table*}
\noindent \textbf{Challenge I: Data Scarcity}. Table~\ref{tab:augmentation} demonstrates the challenge of low-resource vision for existing solutions to data scarcity. Specifically, we test popular data augmentation methods on circuit classification. %
Traditional methods like random crop and flip as well as CutMix~\cite{yun2019cutmix} struggle with our LITE benchmark. %
When using such a small set of images with very fine-grained differences these methods deliver limited additional data diversity. Mixup~\cite{zhang2017mixup} and SyntheticData~\cite{he2022synthetic} obtain better performance as they can create more diverse training data by mixing samples and utilizing generative models. 
Although DA-Fusion~\cite{trabucco2023effective} and SyntheticData~\cite{he2022synthetic} use generative models to obtain more data, they only consider generated images that are similar to the original ones, \ie label-preserving. In contrast, our baseline considers both label-preserving and label-breaking generated images and therefore benefits from many more data points crucial for low-resource vision tasks. This is demonstrated further in the appendix with results for the other two tasks. Combining our baseline solutions to all three challenges results in the best performance, highlighting the multifaceted nature of low-resource vision.

\noindent \textbf{Challenge II: Fine-Grained}. 
We investigate how well recent state-of-the-art fine-grained methods~\cite{bhunia2022adaptive,wang2023open,van2023pdisconet} can tackle the challenge of low-resource vision in  Table~\ref{tab:fine_grained}. We show results for circuit diagram classification here, results for the other two tasks can be found in the appendix. 
We use publicly available implementations except for PLEor~\cite{wang2023open}, which we re-implement ourselves. 
Since fine-grained methods assume there is sufficient data for model learning, they suffer from severe overfitting, degrading the performance of the zero-shot transfer.
We are able to improve performance with our tokenization for fine-grained which attends to fine-grained differences with only a few additional parameters. %

\begin{table}[t!]
\centering
\resizebox{1\linewidth}{!}{
\begin{tabular}{lcc}
\toprule
 & \multicolumn{2}{c}{\textbf{Circuit Classification}} \\
 \cmidrule(lr){2-3} 
 & Top-1 (\%) $\uparrow$ & Top-5 (\%) $\uparrow$ \\%
\midrule
Zero-Shot Transfer & 19.3 & 45.1 \\%
Full-Parameter Finetuning & 13.2 & 38.6 \\
\rowcolor{gray!20}
\multicolumn{3}{l}{\textbf{Transfer Learning}}\\
Linear Probe & 18.7 & 45.9\\%
TOAST~\cite{toast} & 16.4 & 43.3 \\
CLIP-Adapter~\cite{clipadapter} & 16.3 & 42.9 \\%
IA3~\cite{liu2022few} & 18.2 & 45.4\\%
VPT~\cite{VPT} & 19.4 & 45.2 \\%
LoRA~\cite{hu2021lora} & 15.5 & 42.2 \\%
AdaptFormer~\cite{adaptformer} & 19.8 & 45.5\\%
\rowcolor{gray!20}
\multicolumn{3}{l}{\textbf{\textit{Our Baselines}}} \\
Attention for Specialized Domains & \textbf{\hblue{20.6}} &  \textbf{\hblue{47.0}} \\
Combination of Baselines & \textbf{\hred{24.1}} &
\textbf{\hred{49.3}} \\%
\bottomrule
\end{tabular}
}
\vspace{-1em}
\caption{\textbf{Challenge III: Specialized Domain}. \textbf{\hred{Red}} marks the best and \textbf{\hblue{blue}} marks the second. State-of-the-art transfer learning methods focus on common natural images similar to the training data of foundation models, therefore they struggle with low-resource tasks. As a result, our simple baselines can easily lead to improvements. 
}
\vspace{-1em}
\label{tab:transfer}
\end{table}


\noindent \textbf{Challenge III: Specialized Domain}. 
We consider several state-of-the-art transfer learning methods~\cite{toast,clipadapter,liu2022few,VPT,hu2021lora,adaptformer} for adaptation to the specialized domains of our low-resource vision tasks. We show results in Table~\ref{tab:transfer}. All existing baselines struggle to improve over zero-shot transfer with only AdaptFormer giving a slight improvement. %
While more suited to limited data than the fine-grained methods current transfer learning approaches still struggle with the severely limited data of low-resource tasks. They are also not designed to attend to fine-grained details. 
Our attention for specialized domains enables better generalization %
while introducing minimal parameters. Combining all our low-resource baselines to consider all three major challenges further improves the result. However, this is only an initial step towards solving low-resource vision.
%

\begin{figure*}[t!]
\centering
\includegraphics[width=0.8\linewidth]{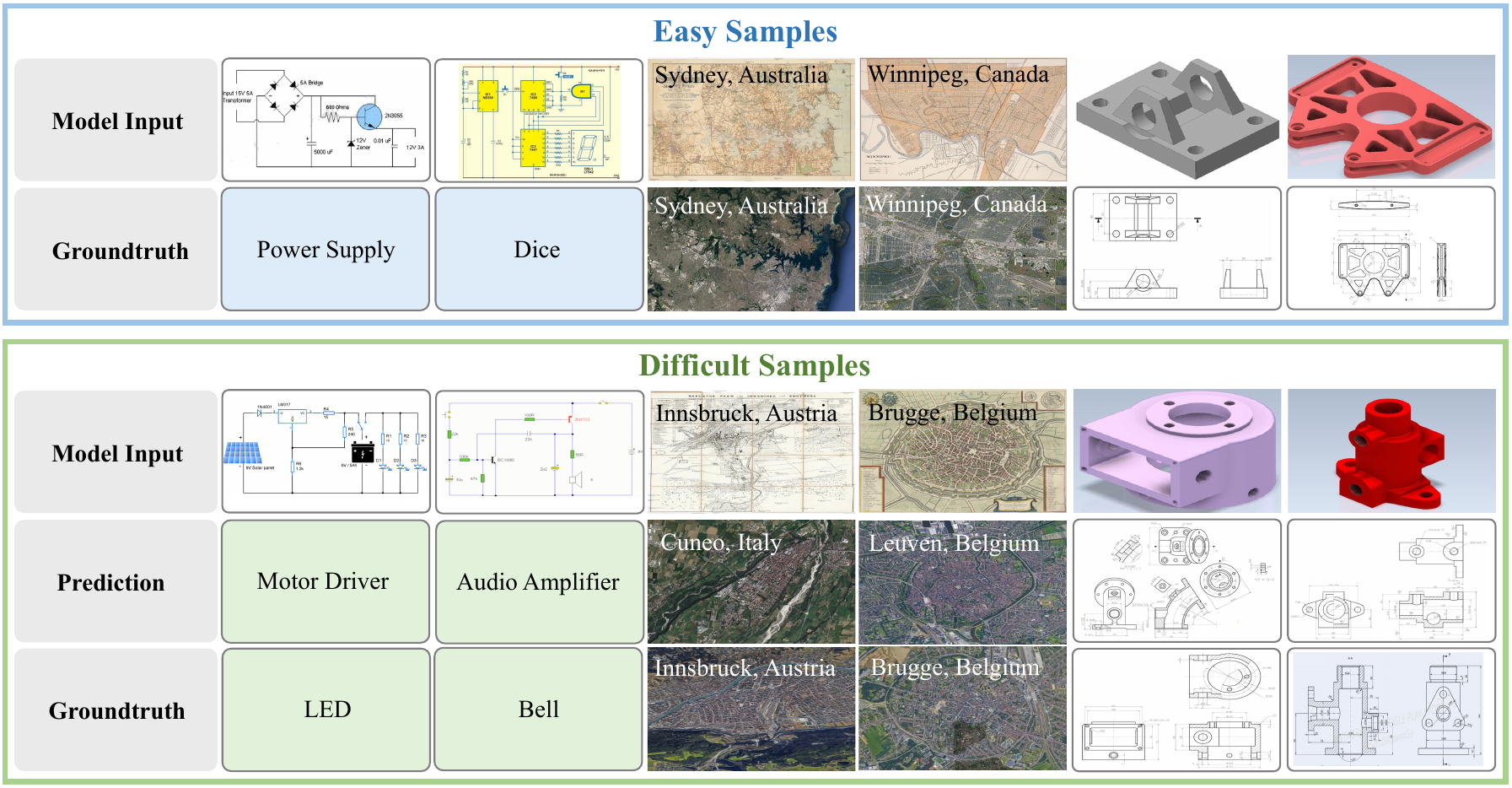}
\vspace{-1em}
\caption{\textbf{Qualitative Results}. We show easy and difficult samples for our baselines. Our baselines can recognize prominent patterns in low-resource data, such as the coastline in the map of Sydney. However, they are overconfident, often basing predictions on one key region such as the presence of the battery in the LED circuit. Our baselines also cannot generalize to rarer image styles such as the Innsbruck map. %
}
\vspace{-1em}
\label{fig:qualitative}
\end{figure*}

\vspace{-0.1em}
\subsection{Our Baselines on Different Foundation Models}
\vspace{-0.3em}

In Table~\ref{tab:foundation_exp}, 
we demonstrate that our low-resource baselines can be plugged into different foundational models by adding them to CLIP~\cite{CLIP}, BLIP~\cite{BLIP}, SAM~\cite{kirillov2023segment}, AIM~\cite{el2024scalable}, DINOv2~\cite{oquab2023dinov2} and ImageBind~\cite{imagebind}. 
All six foundation models can be improved by a large margin with adaptation to low-resource tasks. For example, by adding AdaptFormer, we observe +33.7\% R@1 for CLIP, +40.3\% R@1 for AIM, +40.1\% R@1 for DINOv2 and +41.1\% R@1 for ImageBind on mechanical drawing retrieval. %
The adaptation allows the foundation model features to be better suited to the specific domain of a low-resource task and can thus distinguish images with distinctive patterns. 
Adding our simple baselines results in further improvements. 
For example, there is an additional +4.7\% R@1 improvement for CLIP and DINOv2 and +5.7\% for ImageBind on mechanical drawing retrieval. %
Nevertheless, the performance of our baselines with the best-performing foundation model is still low, \eg, 24.1\% Top-1 accuracy on circuit diagram classification and 37.1 R@1 on historic map retrieval. %
Thus, the proposed low-resource tasks are still far from solved and warrant further study.

\vspace{-0.1em}
\subsection{Discussion}
\vspace{-0.3em}
\noindent \textbf{Qualitative Results}. 
We present qualitative results in Figure~\ref{fig:qualitative}. Our model successfully handles cases where a portion of the image is a clear indication of the label. 
For example, dice circuits contain a digital number display. 
For historic map retrieval, correct examples have a unique coastline or river path, while in mechanical drawings the correct component is clear from all views in the drawing. 
However, our baselines suffer when the relationships between multiple image regions are key. For instance, the horn in the bell circuit diagram also appears in audio amplifiers. The mechanical drawing failure cases appear correct from one drawing perspective but not the others. Our baselines also struggle when the image style is rare in training, as in the  Innsbruck map. 

\noindent\textbf{Opportunities for Future Work.}
While our baselines have made a step towards adapting foundation models to low-resource vision tasks, these tasks are still far from solved. Our baselines still struggle to focus on informative regions due to the unfamiliar specialized domains, the fine-grained details within images, and the limited data we have to adapt foundation models. %
To better tackle the limited data, future works could focus on creating a greater diversity of generated data and explore whether seemingly irrelevant existing data could have some benefit to low-resource tasks. It is also important to consider the relationships between multiple image regions in order to make better fine-grained distinctions. To improve adaptation to specialized domains one possibility is to make the input data more suitable for foundation models with prompt learning or other techniques. Alternatively, future works could consider how foundation models can learn representations that are generalizable to non-natural images. 
In addition to these possible directions, there are also further challenges of low-resource vision beyond the three main challenges this paper explores. For instance, we consider the shift to specialized domains but not the domain shift between the input and ground truth or the sub-domains within the set of images. We also do not explicitly tackle the challenges of huge intra-class variation and imbalanced representation in the limited training set. %
Thus, there is still much room for further work on our low-resource benchmark.

\vspace{-0.3em}
\section{Conclusion}
\vspace{-0.3em}

This paper studies low-resource vision. We collect a benchmark of truly low-resource vision tasks and find these tasks share three challenges: extremely limited data, fine-grained differences between images, and highly specialized domains. To combat these challenges we investigate the generalization capability of foundation models, but find they struggle on low-resource vision tasks. We thus propose three baselines, one per challenge, in a step to solving low-resource vision. 
These baselines improve over prior works tackling individual challenges and can be easily plugged into different foundation models. 
Nevertheless, low-resource vision is still under-explored with many opportunities for future work. 

\medskip

\noindent \textbf{Acknowledgement. }
This work is financially supported by the Inception Institute of Artificial Intelligence, the University of Amsterdam and the allowance Top consortia for Knowledge and Innovation (TKIs) from the Netherlands Ministry of Economic Affairs and Climate Policy.

{
    \small
    \bibliographystyle{ieeenat_fullname}
    \bibliography{main}
}

\clearpage

{
\newpage
   \twocolumn[
    \centering
    \Large
    \textbf{\thetitle}\\
    \vspace{0.5em}Appendix \\
    \vspace{1.0em}

\begin{center}
\centering
\captionsetup{type=table}
\resizebox{0.9\linewidth}{!}{
\begin{tabular}{lcccccccc}
\toprule
 & \multicolumn{2}{c}{\textbf{Circuit Diagram Classification}} & \multicolumn{3}{c}{\textbf{Historic Map Retrieval}} & \multicolumn{3}{c}{\textbf{Mechanical Drawing Retrieval}} \\
 \cmidrule(lr){2-3} \cmidrule(lr){4-6} \cmidrule(lr){7-9} 
 & Top-1 (\%) $\uparrow$ & Top-5 (\%) $\uparrow$ & R@1 $\uparrow$ & R@5 $\uparrow$ & MnR $\downarrow$ & R@1 $\uparrow$ & R@5 $\uparrow$ & MnR $\downarrow$ \\%
\midrule
Zero-Shot Transfer & 19.3 & 45.1 & 28.1 & 62.1 & 10.1 & 13.2 & 26.3 & 83.1 \\
Linear Probe & 18.7 & 45.9 & - & - & - & - & - & -  \\
\rowcolor{gray!20}
\multicolumn{9}{l}{\textbf{ LoRA}~\cite{hu2021lora}}  \\
LoRA & 15.5 & 42.2 & 34.0 & 69.2 & 9.1 & 41.8 & 66.7 & 19.0 \\
+ Generated Data for Data Scarcity & 18.1 & 44.6 & 35.7 & 71.0 & 8.7 & 43.2 & 68.9 & 17.2  \\
+ Tokenization for Fine-Grained & 19.8 & 46.0 & 36.7 & 72.2 & 8.6 & 44.9 & 70.2 & 16.1  \\
+ Attention for Specialized Domains & \textbf{21.0} & \textbf{47.3} & \textbf{37.9} & \textbf{73.3} & \textbf{8.4} & \textbf{46.4} & \textbf{72.3} & \textbf{14.8} \\
\rowcolor{gray!20}
\multicolumn{9}{l}{\textbf{ AdaptFormer}~\cite{adaptformer}}  \\
AdaptFormer & 19.8 & 45.5 & 30.3 & 62.6  & 13.4 & 54.3 & 76.6 & 13.8 \\
+ Generated Data for Data Scarcity & 21.3 & 47.0 & 33.7 & 64.7 & 11.5 & 57.2 & 79.1 & 12.0 \\
+ Tokenization for Fine-Grained & 22.7 & 48.1 & 34.9 & 66.4 & 10.8 & 58.8 & 81.2 & 11.4 \\
+ Attention for Specialized Domains & \textbf{24.1} & \textbf{49.3} & \textbf{36.4} & \textbf{68.0} & \textbf{9.8} & \textbf{60.0} & \textbf{82.5} & \textbf{10.2} \\
\bottomrule
\end{tabular}
}
\vspace{-0.5em}
\captionof{table}{\textbf{Combination of Our Baselines}. Our generated data for data scarcity can mitigate the overfitting considerably for both LoRA and AdaptFormer. The attention for specialized domains and tokenization for fine-grained further contribute to performance improvement by helping the model focus on task-relevant regions and fine-grained details. 
}
\label{tab:ablation}
\end{center}


   ] 
}

 \begin{table*}[t!]
 \centering
\captionsetup{type=table}
\resizebox{0.9\linewidth}{!}{
\begin{tabular}{lcccccccc}
\toprule
 & \multicolumn{2}{c}{\textbf{Circuit Diagram Classification}} & \multicolumn{3}{c}{\textbf{Historic Map Retrieval}} & \multicolumn{3}{c}{\textbf{Mechanical Drawing Retrieval}} \\
 \cmidrule(lr){2-3} \cmidrule(lr){4-6} \cmidrule(lr){7-9} 
 & Top-1 (\%) $\uparrow$ & Top-5 (\%) $\uparrow$ & R@1 $\uparrow$ & R@5 $\uparrow$ & MnR $\downarrow$ & R@1 $\uparrow$ & R@5 $\uparrow$ & MnR $\downarrow$ \\ %
\midrule
Zero-Shot Transfer & 19.3 & 45.1 & 28.1 & 62.1 & 10.1 & 13.2 & 26.3 & 83.1 \\
\rowcolor{gray!20}
\multicolumn{9}{l}{\textbf{Simple Transformations}}  \\
Random Cropping + Random Flipping & 19.8 & 45.3 & 30.3 & 62.6 & 13.4 & 54.3 & 76.6 & 13.8 \\
Mixup~\cite{zhang2017mixup} & 20.8 & 46.0 & 28.4 & 55.7 & 15.2 & 50.5 & 72.9 & 14.9 \\
CutMix~\cite{yun2019cutmix} & 20.0 & 45.5 & 28.4 & 62.6 & 12.5 & 54.9 & 78.1 & 13.1 \\
Random Erasing~\cite{zhong2020random} & 20.8 & 46.2 & 23.5 & 49.6 & 17.1  & 54.6 & 78.0 & 13.2\\
\rowcolor{gray!20}
\multicolumn{9}{l}{\textbf{Generative Models}}  \\
DA-Fusion~\cite{trabucco2023effective} & 19.6 & 45.1 & 29.8 & 61.3 & 14.7 & 54.5 & 77.6 & 13.5 \\ %
SyntheticData~\cite{he2022synthetic} & 20.8 & 46.0 & 30.4 & 62.9 & 12.6 & 55.9 & 78.9 & 13.0 \\ %
\rowcolor{gray!20}
\multicolumn{9}{l}{\textbf{\textit{Our Baselines}}}  \\
Generated Data for Data Scarcity & \textbf{\hblue{21.3}} & \textbf{\hblue{46.9}} & \textbf{\hblue{33.7}} & \textbf{\hblue{64.7}} & \textbf{\hblue{11.5}} & \textbf{\hblue{57.2}} & \textbf{\hblue{79.1}} &  \textbf{\hblue{12.0}} \\
Combination of Baselines & \textbf{\hred{24.1}} & \textbf{\hred{49.3}} & \textbf{\hred{36.4}} & \textbf{\hred{68.0}} &  \textbf{\hred{9.8}} & \textbf{\hred{60.0}} & \textbf{\hred{82.5}} & \textbf{\hred{10.2}} \\
\bottomrule
\end{tabular}
}
\vspace{-1em}
\captionof{table}{\textbf{Challenge I: Data Scarcity}. We mark the best in \textbf{\hred{red}} and the second in \textbf{\hblue{blue}}. Simple transformations do little to improve the diversity of training data. We obtain the best data diversity and thus the best baseline performance on all the three low-resource tasks with our baselines which leverage both similar and dissimilar images produced by generative models. 
}
 \vspace{-0.5em}
\label{tab:augmentation_full}
\end{table*}

\begin{table*}[t!]
\centering
\resizebox{0.85\linewidth}{!}{
\begin{tabular}{lcccccccc}
\toprule
 & \multicolumn{2}{c}{\textbf{Circuit Diagram Classification}} & \multicolumn{3}{c}{\textbf{Historic Map Retrieval}} & \multicolumn{3}{c}{\textbf{Mechanical Drawing Retrieval}} \\
 \cmidrule(lr){2-3} \cmidrule(lr){4-6} \cmidrule(lr){7-9} 
 & Top-1 (\%) $\uparrow$ & Top-5 (\%) $\uparrow$ & R@1 $\uparrow$ & R@5 $\uparrow$ & MnR $\downarrow$ & R@1 $\uparrow$ & R@5 $\uparrow$ & MnR $\downarrow$ \\
\midrule
Zero-Shot Transfer & 19.3 & 45.1 & 28.1 & 62.1 & 10.1 & 13.2 & 26.3  & 83.1 \\
\rowcolor{gray!20}
\multicolumn{9}{l}{\textbf{Fine-Grained}}  \\
Adaptive-FGSBIR~\cite{bhunia2022adaptive} & 16.7 & 43.2 & 5.3 & 12.6 & 55.1 & 5.6 & 17.1 & 70.2 \\ 
PLEor~\cite{wang2023open} & 17.1 & 44.1 & 4.6 & 11.8 & 56.5 & 5.0 & 16.9 & 71.3 \\ 
PDiscoNet~\cite{van2023pdisconet} & 16.2 & 43.5 & 5.8 & 13.0 & 53.2 & 5.4 & 17.2 & 69.8 \\ 
\rowcolor{gray!20}
\multicolumn{9}{l}{\textbf{\textit{Our Baselines}}}  \\
Tokenization for Fine-Grained & \textbf{\hblue{20.9}} & \textbf{\hblue{45.5}} & \textbf{\hblue{32.1}} & \textbf{\hblue{64.0}} & \textbf{\hblue{12.3}} & \textbf{\hblue{55.9}} & \textbf{\hblue{78.4}} & \textbf{\hblue{12.7}} \\
Combination of Baselines & \textbf{\hred{24.1}} & \textbf{\hred{49.3}} & \textbf{\hred{36.4}} & \textbf{\hred{68.0}} & \textbf{\hred{9.8}} & \textbf{\hred{60.0}} & \textbf{\hred{82.5}} & \textbf{\hred{10.2}} \\
\bottomrule
\end{tabular}
}
\vspace{-1em}
\caption{\textbf{Challenge II: Fine-Grained}. We mark the best in \textbf{\hred{red}} and the second in \textbf{\hblue{blue}}. Fine-grained recognition methods need thousands of images for model learning, making them unsuited to low-resource tasks. Our tokenization baseline better utilizes the limited training data and makes improvements on all the three low-resource tasks. However, there is much potential for further improvements. 
}
 \vspace{-0.5em}
\label{tab:fine_grained_full}
\end{table*}

\begin{table*}[t!]
\centering
\resizebox{1\linewidth}{!}{
\begin{tabular}{lcccccccccc}
\toprule
 & \multicolumn{2}{c}{\textbf{Circuit Diagram Classification}} & \multicolumn{3}{c}{\textbf{Historic Map Retrieval}} & \multicolumn{3}{c}{\textbf{Mechanical Drawing Retrieval}} & \\
 \cmidrule(lr){2-3} \cmidrule(lr){4-6} \cmidrule(lr){7-9} 
 & Top-1 (\%) $\uparrow$ & Top-5 (\%) $\uparrow$ & R@1 $\uparrow$ & R@5 $\uparrow$ & MnR $\downarrow$ & R@1 $\uparrow$ & R@5 $\uparrow$ & MnR $\downarrow$ & GFLOPs & Params (M)\\
\midrule
Zero-Shot Transfer & 19.3 & 45.1 & 28.1 & 62.1 & 10.1 & 13.2 & 26.3 & 83.1 & 224.6 & 63.3 \\
\rowcolor{gray!20}
\multicolumn{11}{l}{\textbf{\textit{Transfer Learning}}} \\
TOAST~\cite{toast} & 16.4 & 43.3 & 4.2 & 11.5 & 52.4 & 4.4 & 16.3 & 69.0 & 476.2 & 73.8\\
CLIP-Adapter~\cite{clipadapter}  & 16.3 & 42.9 & 3.2 & 15.9 & 42.1 & 7.7 & 23.2 & 59.5  & 224.8 & 64.2 \\ %
IA3~\cite{liu2022few} & 18.2 & 45.4 & 29.1 & 51.3 & 19.5 & 52.0 & 76.7 & 12.7 & 224.6 & 63.6 \\ %
VPT~\cite{VPT} & 19.4 & 45.2 & 36.2 & 61.6 & 13.3 & 47.7 & 72.4 & 13.5 & 233.3 & 63.8 \\
\rowcolor{gray!20}
\multicolumn{11}{l}{\textbf{\textit{LoRA w/ Our Baselines}}} \\
LoRA~\cite{hu2021lora} & 15.5 & 42.2 & 34.0 & 69.2 & 9.1 & 41.8 & 66.7 & 19.0 & 224.7 & 63.7 \\ %
+ Attention for Specialized Domains & 16.9 & 44.5 & 35.1 & \textbf{\hblue{70.9}} & \textbf{\hblue{8.8}} & 43.0 & 69.2 & 17.8 & 224.7 & 63.7 \\
+ Combination of Baselines & \textbf{\hblue{21.0}} & \textbf{\hblue{47.3}} & \textbf{\hred{37.9}} & \textbf{\hred{73.3}} & \textbf{\hred{8.4}} & 46.4 & 72.3 & 14.8 &  233.7 & 63.7 \\
\rowcolor{gray!20}
\multicolumn{11}{l}{\textbf{\textit{AdaptFormer w/ Our Baselines}}} \\
AdaptFormer~\cite{adaptformer}  & 19.8 & 45.5 & 30.3 & 62.6 & 13.4 & 54.3 & 76.6 & 13.8 & 224.6 & 63.4 \\%
+ Attention for Specialized Domains & 20.6 & 47.0 & 31.9 & 64.2 & 12.1 & \textbf{\hblue{56.4}} & \textbf{\hblue{78.3}} & \textbf{\hblue{12.8}} & 224.6 & 63.4 \\
+ Combination of Baselines & \textbf{\hred{24.1}} & \textbf{\hred{49.3}} & \textbf{\hblue{36.4}} & 68.0 & 9.8 & \textbf{\hred{60.0}} & \textbf{\hred{82.5}} & \textbf{\hred{10.2}} & 233.6 & 63.4 \\%
\bottomrule
\end{tabular}
}
\vspace{-1em}
\caption{\textbf{Challenge III: Specialized Domain}. We mark the best in \textbf{\hred{red}} and the second in \textbf{\hblue{blue}}. State-of-the-art transfer learning methods focus on common natural images similar to the training data of foundation models, therefore they struggle with low-resource tasks. As a result, our simple baselines can easily lead to improvements on all the three low-resource tasks. 
}
\label{tab:transfer_full}
\end{table*}


\begin{table*}
\resizebox{1\linewidth}{!}{
\begin{tabular}{cccccccccc}
\toprule
& & \multicolumn{2}{c}{\textbf{Circuit Diagram Classification}} & \multicolumn{3}{c}{\textbf{Historic Map Retrieval}} & \multicolumn{3}{c}{\textbf{Mechanical Drawing Retrieval}} \\
\cmidrule(lr){3-4}  \cmidrule(lr){5-7} \cmidrule(lr){8-10}
\textbf{Label-Preserving} & \textbf{Label-Breaking} & Top-1 (\%) $\uparrow$ & Top-5 (\%) $\uparrow$ & R@1 $\uparrow$ & R@5 $\uparrow$ & MnR $\downarrow$ & R@1 $\uparrow$ & R@5 $\uparrow$ & MnR $\downarrow$ \\
\midrule
  &  & 19.8 & 45.5 & 30.3 & 62.6  & 13.4 & 54.3 & 76.6 & 13.8 \\
 $\checkmark$ & & 20.8 & 46.1 & 32.1 & 63.8 & 12.1 & 56.4 & 78.0 &  12.8 \\
 & $\checkmark$ & 20.4 & 46.0 & 32.4 & 63.5 & 12.5 & 55.7 & 77.5 & 13.0 \\
 $\checkmark$ & $\checkmark$  & \textbf{21.3} & \textbf{46.9} & \textbf{33.7} & \textbf{64.7} & \textbf{11.5} & \textbf{57.2} & \textbf{79.1} & \textbf{12.0} \\
\bottomrule
\end{tabular}
}
\vspace{-1em}
\caption{\textbf{Ablation of Generated Data}. Both label-preserving and label-breaking generated images add more data points into the training data for model learning. Thus, both types of augmentation contribute to the reduction in overfitting. 
}
 \vspace{-0.5em}
\label{tab:ablation_generated_appendix}
\end{table*}

\begin{figure*}[t]
\begin{minipage}{0.3\textwidth}
\includegraphics[width=0.95\linewidth]{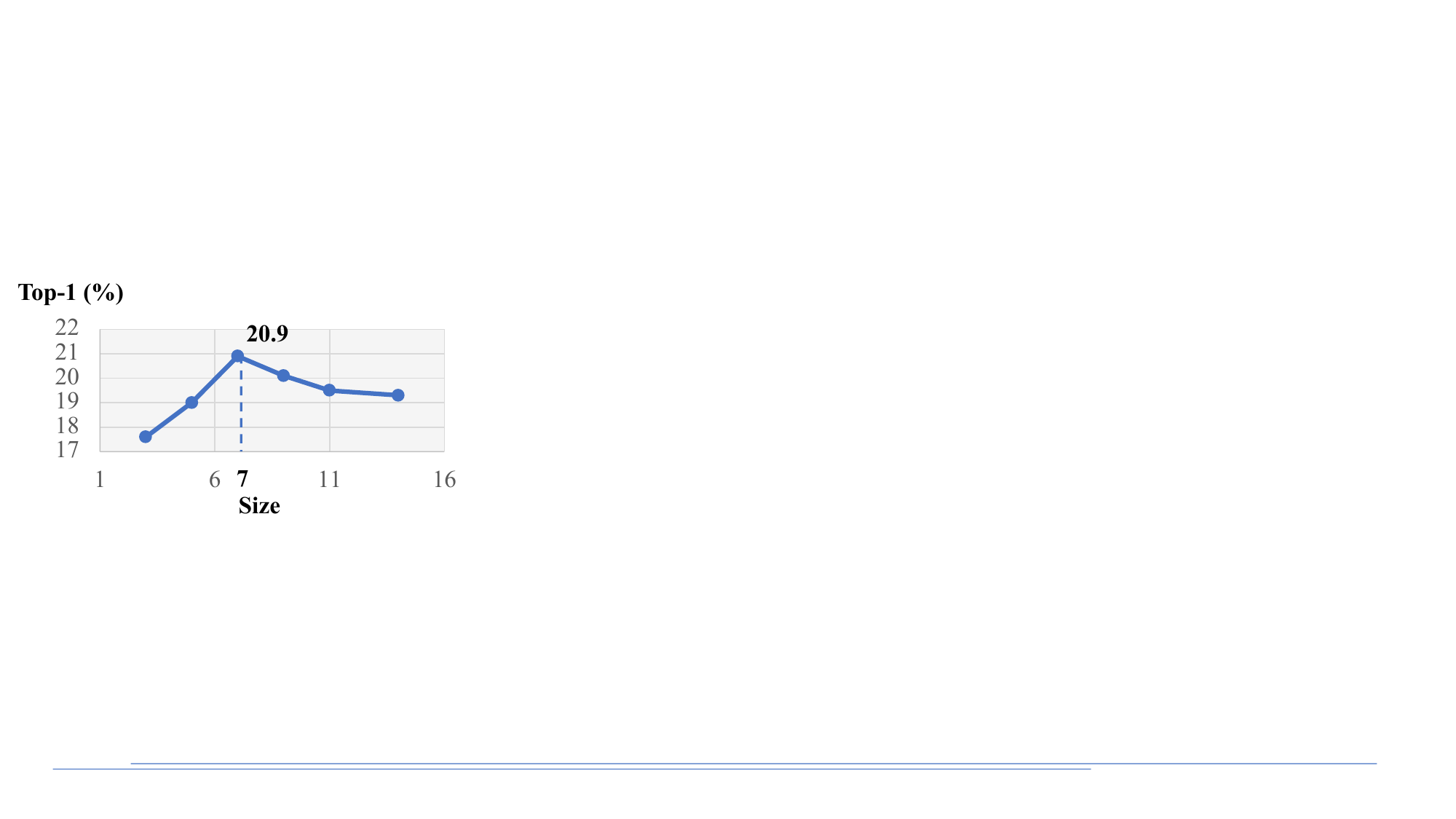}
	 \caption{\textbf{Effect of Sub-Kernel Size} in tokenization for fine-grained. A medium sub-kernel gives a good trade-off between meaningful regions and fine-grained details. 
 }
\label{fig:appendix_sub_kernel_size}
\end{minipage}
\hspace{0.03\textwidth}
\begin{minipage}{0.32\textwidth}
\includegraphics[width=1\linewidth]{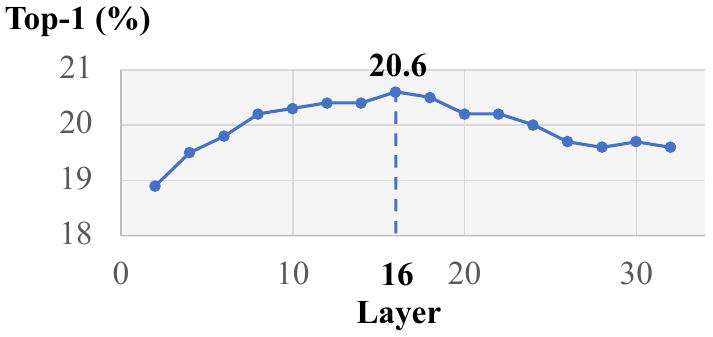}
	\caption{\textbf{Position of Attention} for specialized domains. The attention is reasonably robust to the choice of layer, although the middle layers reach a good trade-off between low- and high-level features.  }
 \label{fig:appendix_att_position}
\end{minipage}
\hspace{0.03\textwidth}
\begin{minipage}{0.3\textwidth}
\includegraphics[width=0.9\linewidth]{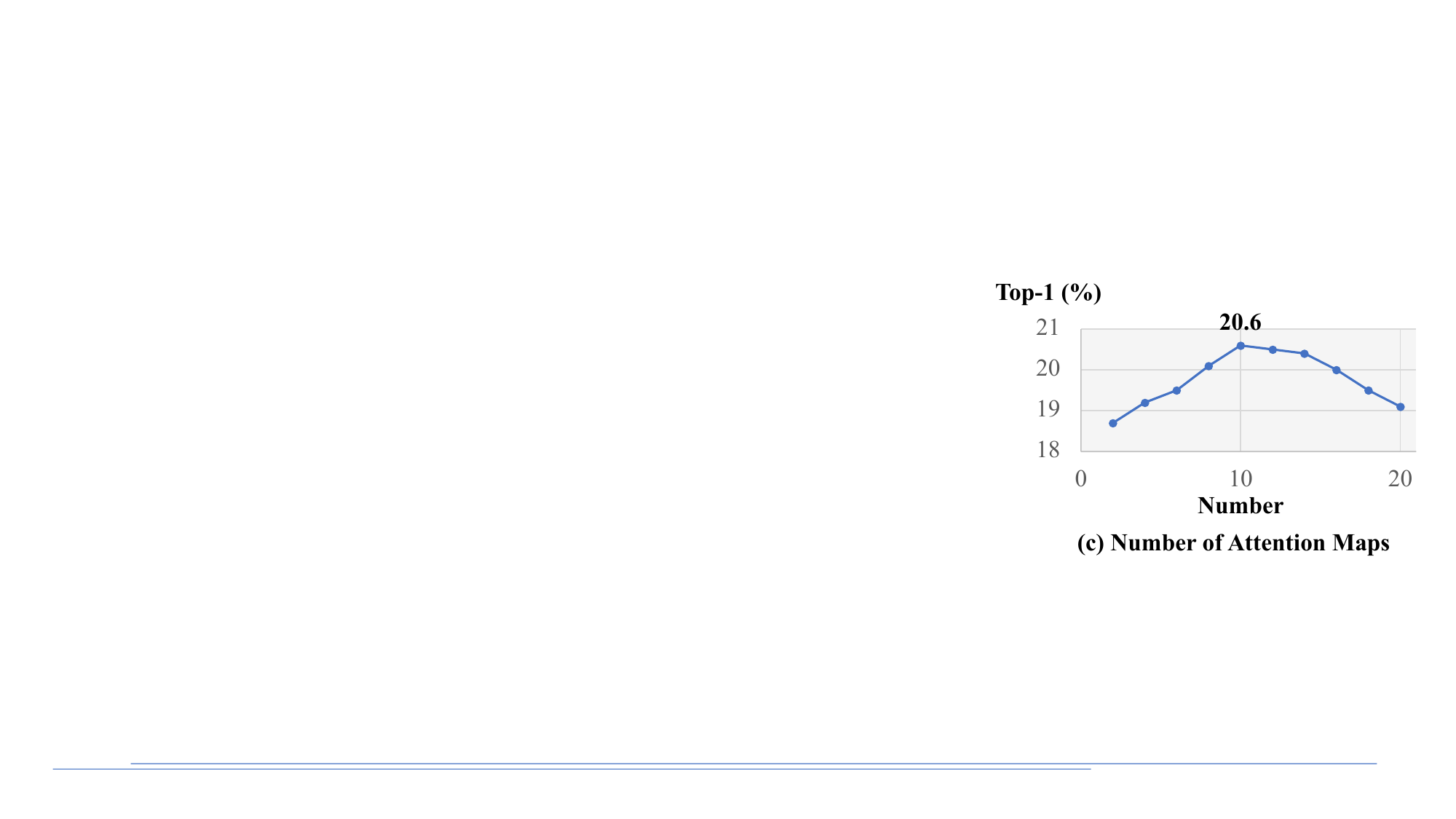}
	\caption{\textbf{Number of Attention Maps} for specialized domains. Learning anywhere from $8$ to $14$ attention maps allows the model to specialize to the domain while avoiding overfitting}
 \label{fig:appendix_att_number}
\end{minipage}
 \vspace{-1.5mm}
\end{figure*}

\begin{figure*}[t!]
\centering
\includegraphics[width=0.6\linewidth]{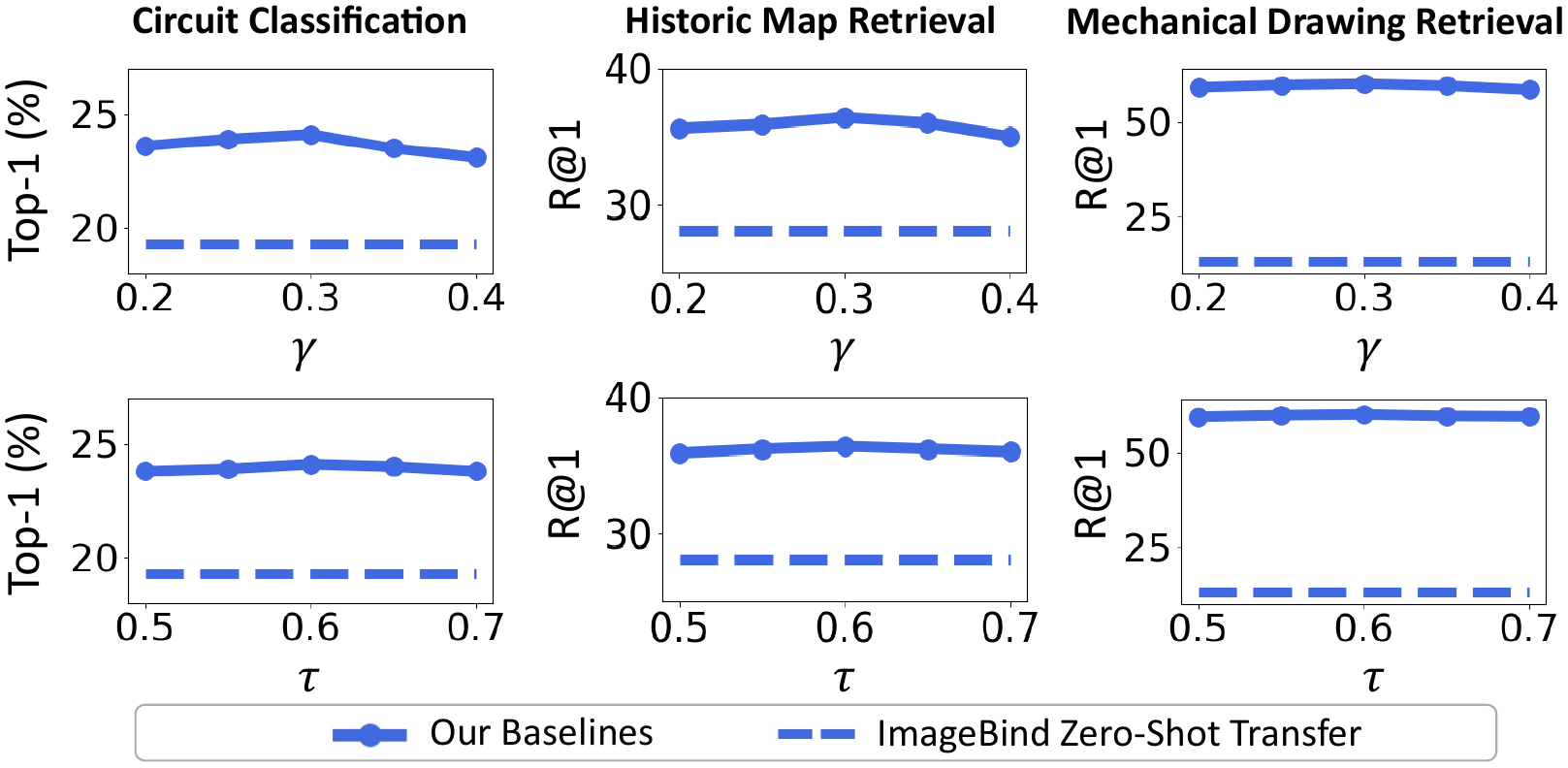}
\vspace{-0.5em}
\caption{\textbf{Diffusion Model Thresholds} have limited influence on baseline I for synthesizing data.}
\label{fig:supp_gamma}
\vspace{-0.5em}
\end{figure*}

\setcounter{section}{0}

\renewcommand{\thesection}{\Alph{section}}
\renewcommand{\thesubsection}{\Alph{section}.\arabic{subsection}}

\begin{figure}[t!]
\centering
\includegraphics[width=1\linewidth]{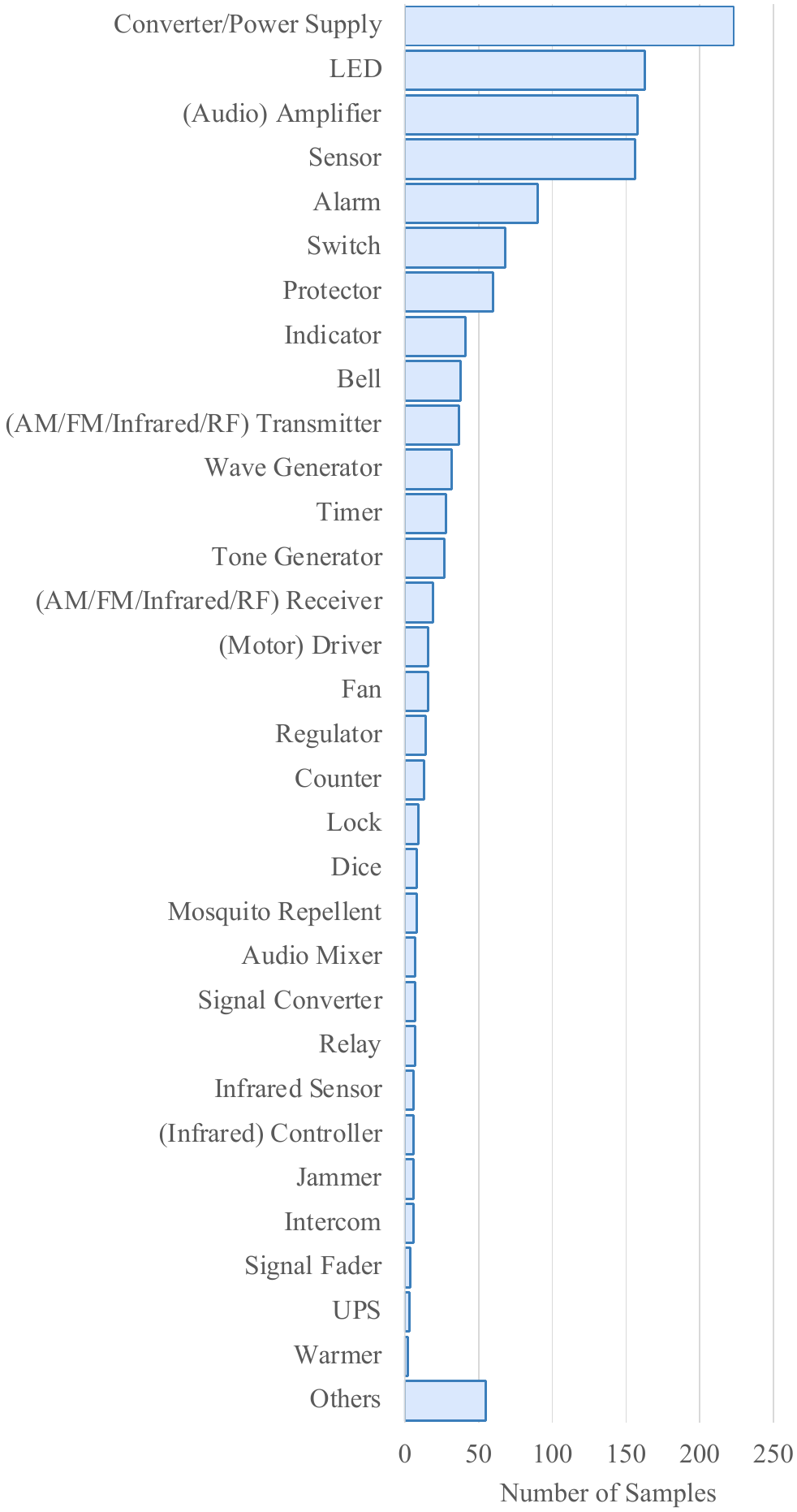}
\vspace{-1.5em}
\caption{\textbf{Class Distribution of Circuit Diagrams}. While power supply, LED, amplifier and sensor have the most samples, it is hard to collect many circuit diagrams for most classes. 
}
\vspace{-1em}
\label{fig:appendix_circuit_distribution}
\end{figure}

\begin{figure}[t!]
\centering
\includegraphics[width=0.8\linewidth]{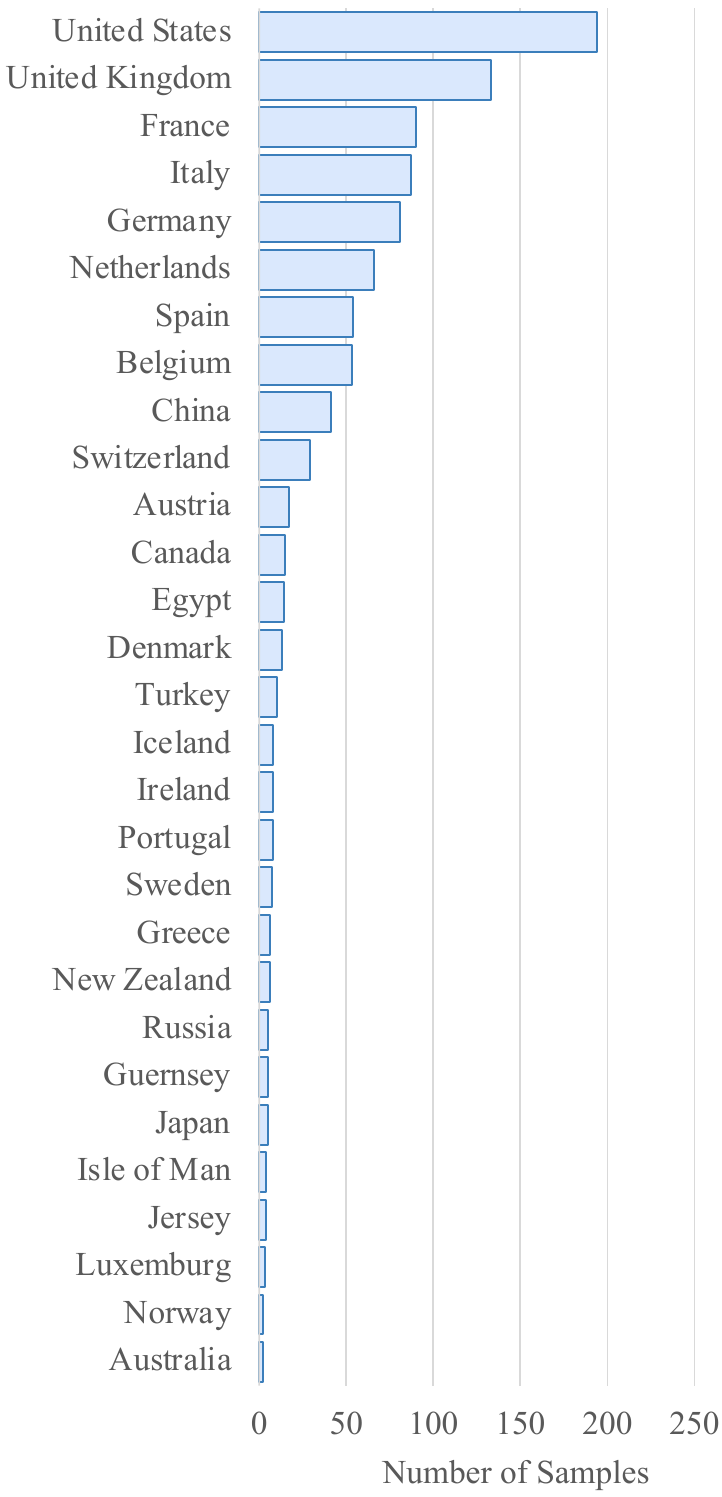}
\vspace{-0.5em}
\caption{\textbf{Country Distribution of Historic Maps}. While we find Europe and United States have the most historic maps online, other regions have much fewer maps available. 
}
\vspace{-0.5em}
\label{fig:appendix_map_distribution}
\end{figure}

\section{Combination of Our Baselines}
\label{sec:appendix_ablation}
\vspace{-0.3em}
\textbf{Method. }
We adapt the foundation model ImageBind by an existing transfer learning method, and add our proposed baselines for the three challenges defined in Section~\ref{sec:LITE}. 
We keep the foundation model parameters frozen and only update the parameters introduced by the transfer learning method and our baselines. 
Our baselines work independently of each other, focusing on different areas of the foundation model: input, tokenization, and attention. Thus, they can be easily combined. 
Specifically, our generated data for data scarcity produce more samples for model learning, while our tokenization for fine-grained better encodes the input image patches into feature tokens. 
As the transfer learning method learns additional parameters inside the foundation model for the low-resource tasks, our attention for specialized domains is inserted into one layer to help the model focus on task-relevant regions.

\noindent \textbf{Results. }
In Table~\ref{tab:ablation}, we ablate the effect of our three proposed baselines (in Section~\ref{sec:challenges}), \ie, generated data for data scarcity, tokenization for fine-grained and attention for specialized domains. 
We consider both LoRA~\cite{hu2021lora} and AdaptFormer~\cite{adaptformer} as the additional transfer learning parameters.
By adding our generated data, the overfitting issue from limited training data can be alleviated considerably. For instance, this gives +2.6\% Top-1 accuracy on circuit diagram classification with LoRA and 3.4\% in R@1 on historic maps with AdaptFormer. 
This is because the label-breaking images enlarge the data space to help the representation learning. 
With our tokenization for fine-grained, the performance is further improved by +1.7\% Top-1 with LoRA and +1.4\% with AdaptFormer on circuit diagram classification. 
This demonstrates the benefit of processing smaller regions compared to the original kernel so that the fine-grained details can be discovered. 
Adding attention for specialized domains to combat the out-of-distribution challenge delivers a further +1.2\% Top-1 with LoRA and 1.4\% with Adaptformer on circuit classification. 
Historic map retrieval and mechanical drawing retrieval obtain similar improvements. 
Our baselines are effective additions to both LoRA and AdaptFormer, increasing the results on circuit classification by 5.5\% Top-1 and 4.3\% respectively, with similar improvements for historic map retrieval and mechanical drawing retrieval. 

\vspace{-0.5em}
\section{Challenge Results on All Tasks}
\vspace{-0.3em}

In Section~\ref{sec:challenge_results} in the main paper, we demonstrate the challenges of low-resource vision for existing solutions on circuit diagram classification. Here, we present the same experiments on all three low-resource tasks including historic map and mechanical drawing retrieval. 

\noindent \textbf{Challenge I: Data Scarcity}. 
In Table~\ref{tab:augmentation_full}, we show the challenge of low-resource vision for existing solutions to data scarcity. Existing approaches are effective for mechanical drawing retrieval but give little improvement over zero-shot transfer on circuit diagram classification and historic map retrieval.
Our generated data for data scarcity benefits all three tasks, giving the most improvement on historic map retrieval and mechanical drawing retrieval. 
This is because the domain gap between natural images and the data of these two tasks is slightly smaller making the label-breaking augmentations look more realistic (we show visualizations in Section~\ref{sec:appendix_visualization}). Our generated data for data scarcity thus increases the data diversity more effectively for historic map and mechanical drawing retrieval. 
While our combination of baselines does deliver improvements on all three tasks, they are still far from solved, uncovering the difficulties of low-resource vision. 

\noindent \textbf{Challenge II: Fine-Grained}. 
We investigate how well recent state-of-the-art fine-grained methods~\cite{bhunia2022adaptive,wang2023open,van2023pdisconet} can tackle the challenge of low-resource vision in Table~\ref{tab:fine_grained_full}. 
While existing fine-grained methods assume there is sufficient data for model learning, they suffer from severe overfitting. This is demonstrated best on historic map and mechanical drawing retrieval where the results of existing fine-grained methods are much lower than zero-shot transfer. 
In contrast, our baseline for fine-grained attends to fine-grained differences with only a few additional parameters so that the performance can be improved on all three tasks.

\noindent \textbf{Challenge III: Specialized Domain}. 
We consider several state-of-the-art transfer learning methods~\cite{toast,clipadapter,liu2022few,VPT,hu2021lora,adaptformer} for adaptation to the specialized domains of our low-resource vision tasks. We show results in Table~\ref{tab:transfer_full}. 
Since our baselines can be used in combination with any transfer learning method we plug them into two such methods: LoRA~\cite{hu2021lora} and AdaptFormer~\cite{adaptformer}. 
While AdaptFormer performs better than other transfer learning methods on circuit diagram classification and mechanical drawing retrieval, LoRA is favorable on historic map retrieval.  
With our baselines, the results are improved further without much computation burden or parameters added. 
However, there is still no single model that always surpasses the others on all three low-resource tasks. 
Thus, this is only an initial step towards solving the challenges of low-resource vision.

\vspace{-0.3em}
\section{Ablations and Hyperparameter Analysis}
\vspace{-0.3em}

\subsection{Generated Data for Data Scarcity}
\label{sec:appendix_generated_ablation}

\noindent \textbf{Effect of Label-Preserving and Label-Breaking Images}. 
Our generated data for data scarcity in Section~\ref{sec:diverse} use two types of augmentations for model learning, \ie, label-preserving and label-breaking. 
While we adopt a supervised learning objective for label-preserving images with groundtruth labels, a self-supervised contrastive learning objective is applied to label-breaking images. 
In Table~\ref{tab:ablation_generated_appendix}, we ablate their effect on our low-resource benchmark. 
Both types of images improve the diversity of training data. 
Thus, they reduce the overfitting and improve the model adaptation to our low-resource settings considerably. 
We provide visualizations of these two types of augmentations in Section~\ref{sec:appendix_visualization}.

\noindent \textbf{Diffusion Model Threshold}. Our generated data baseline is insensitive to the selection of $\gamma$ and $\tau$ thresholds and improves over ImageBind zero-shot transfer for all thresholds tested in Figure~\ref{fig:supp_gamma}.

\subsection{Tokenization for Fine-Grained}
\label{sec:appendix_tokenization_ablation}

\noindent \textbf{Effect of Sub-Kernel Size}. In Section~\ref{sec:selective}, we introduce our tokenization for fine-grained baseline, where the original kernel for linear projection is divided into sub-kernels for encoding the input image patches. 
As the sub-kernels have smaller receptive fields, more simple patterns are encoded from the smaller input image patches. 
Here, we ablate the effect of sub-kernel size in Figure~\ref{fig:appendix_sub_kernel_size} on circuit diagram classification. 
While small sub-kernel sizes cannot cover meaningful regions and therefore only extract the texture information, large sub-kernel sizes focus more on global information 
than fine-grained details. Thus, adopting a medium size, \ie, $7 {\times} 7$, delivers the best performance.

\begin{figure*}[t!]
\centering
\includegraphics[width=1\linewidth]{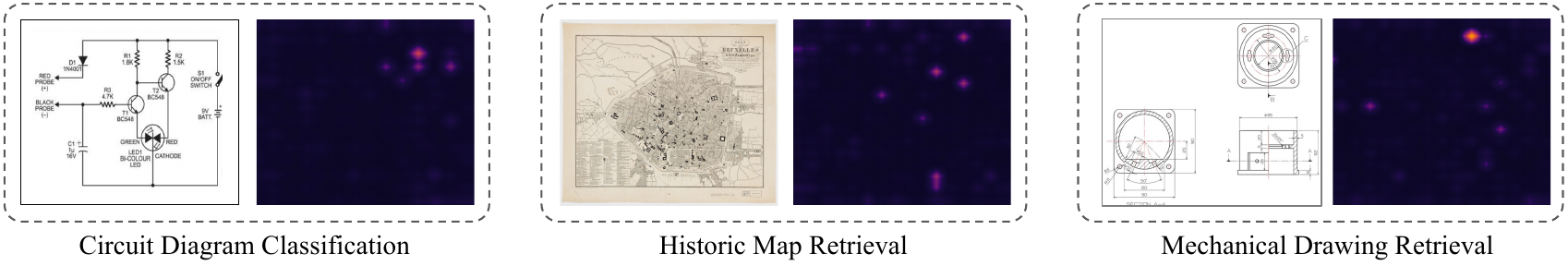}
\vspace{-2em}
\caption{\textbf{Attention Maps from Vision Foundation Model}. 
We show the highest activation on each region across all the attention maps of the middle transformer block. 
Only a few regions are activated for the three images. This means that the vision foundation model fails to understand the interaction between different image regions. Thus, vision foundation models need proper adaptation for low-resource tasks. 
}
\label{fig:appendix_att_foundation}
\end{figure*}

\begin{figure*}[t!]
\centering
\includegraphics[width=1\linewidth]{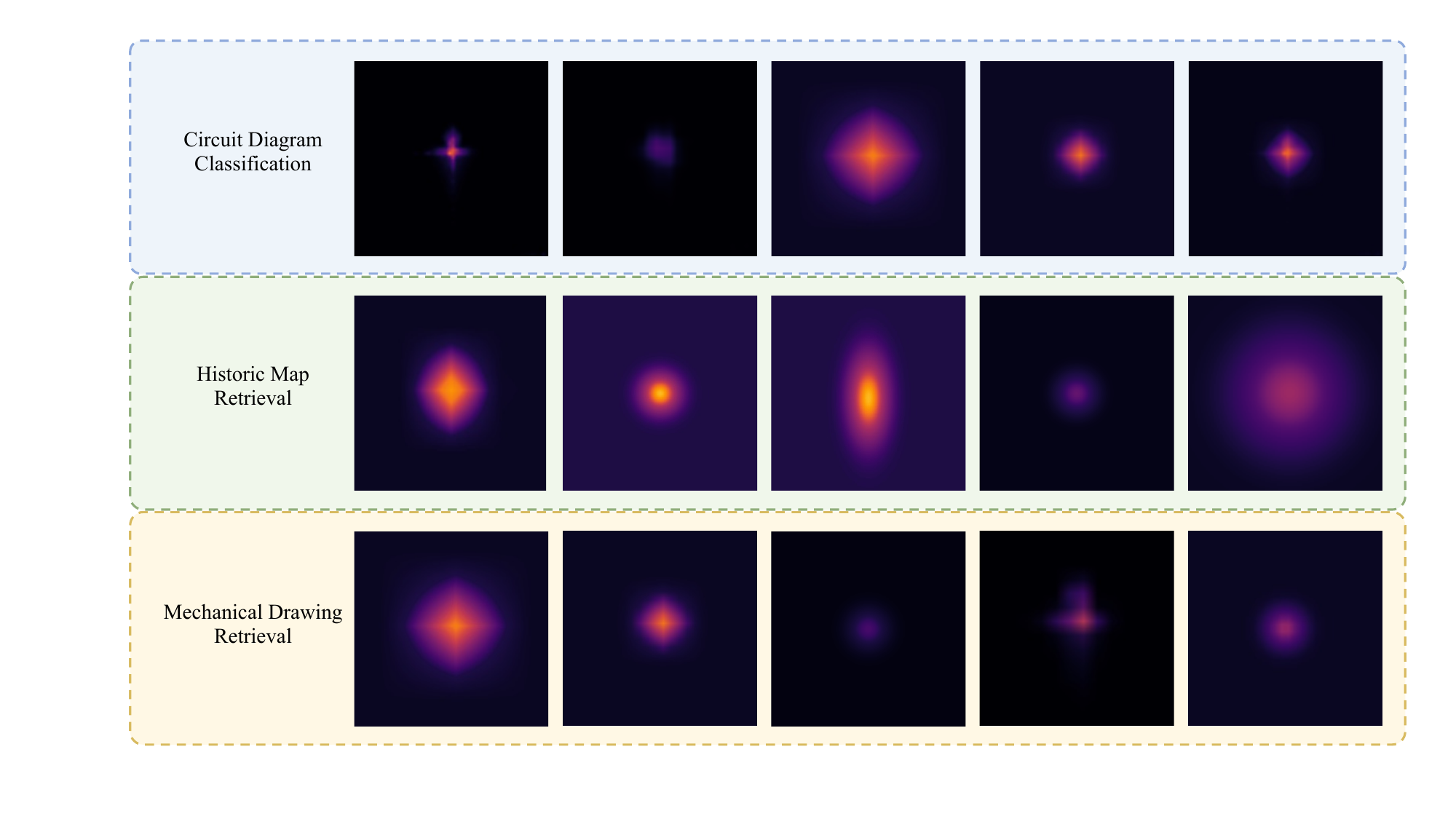}
\vspace{-2em}
\caption{\textbf{Attention for Specialized Domains}. While the attention maps for circuit diagram classification and mechanical drawing retrieval focus more on vertical and horizontal regions, those for historic maps highlight different local regions and tend to have much larger `receptive fields' than the other two tasks. 
}
\vspace{-1em}
\label{fig:appendix_att_visualization}
\end{figure*}

\subsection{Attention for Specialized Domains}
\label{sec:appendix_attention_ablation}

In Section~\ref{sec:adaptive}, we introduce our attention for specialized domains. Here, we discuss the effect of its position as well as the number of maps. 

\noindent \textbf{Position of Attention for Specialized Domains}. 
We add our attention for specialized domains into only one layer to avoid introducing many additional parameters, which can result in overfitting. 
In Figure~\ref{fig:appendix_att_position} we measure the effect of adding our attention to different transformer layers in circuit classification. %
As shallow layers focus on low-level, simple patterns, deep layers extract semantic features. The middle layers reach a trade-off between low- and high-level features. 
Nonetheless, our attention for specialized domains is reasonably robust to the choice of layer.

\noindent \textbf{Number of Attention Maps}. Using attention for specialized domains in the middle layer, we further study the effect of the number of attention maps $C$ %
on circuit diagram classification in Figure~\ref{fig:appendix_att_number}. Using as few as two attention maps doesn't allow the model to fully specialize to the low-resource domain, while with $20$ maps, the model overfits to the training data and cannot generalize to the variations seen at inference. %
Learning $10$ maps reaches the best trade-off, although the model is beneficial with any choice between 8 and 14.

\begin{figure*}[t!]
\centering
\includegraphics[width=0.8\linewidth]{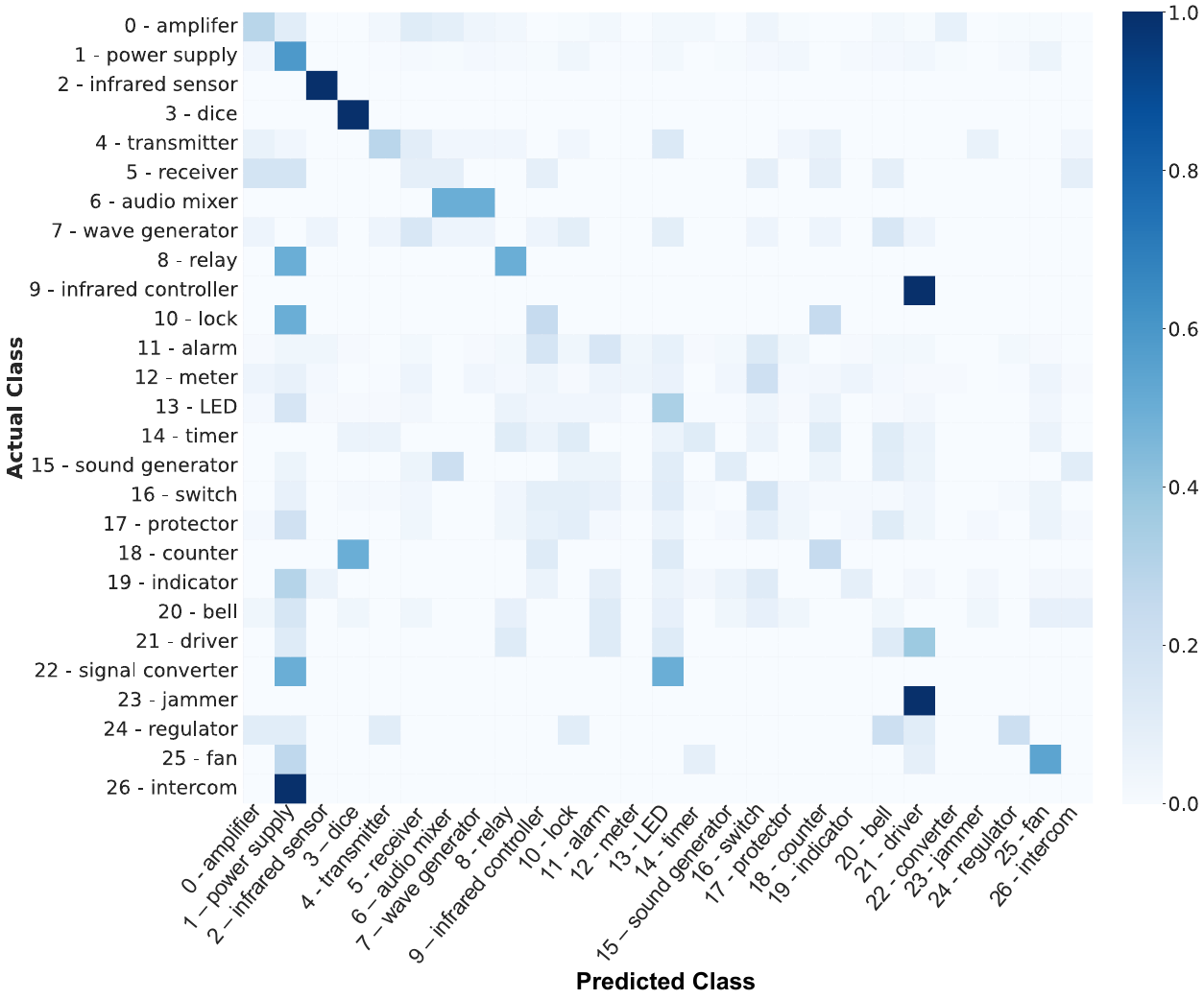}
\vspace{-0.8em}
\caption{\textbf{\textbf{Confusion Matrix}} reveals our baselines achieve stronger performance on classes with prominent patterns. }
\label{fig:supp_confusion}
\vspace{-0.5em}
\end{figure*}

\section{Low-Resource Image Transfer Evaluation}

In Section~\ref{sec:LITE} in the main paper, we introduce our Low-Resource Image Transfer Evaluation benchmark. Here, we provide more details about our three low-resource tasks. 

\noindent \textbf{Task I: Circuit Diagram Classification}. 
We collect 297 circuit diagrams from~\cite{book}, 175 from Gadgetronicx~\cite{Gadgetronicx} and 860 from Circuit Digest~\cite{circuitdigest}. 
This results in a total of 1,332 circuit diagrams. 
We divide them into 32 classes and present the class distribution in Figure~\ref{fig:appendix_circuit_distribution}. 
The power supply contains 223 samples, which is the most among all classes. 
We also find many samples for LED with 163 diagrams, amplifier with 158, and sensor with 156. 
However, we also find it hard to collect circuit diagrams for many classes, \eg, relay, jammer, intercom, and signal fader. 
For instance, we can only get 7 samples of relays and 6 depicting jammers.

\noindent \textbf{Task II: Historic Map Retrieval}. 
All the historic maps come from OLD MAPS ONLINE~\cite{oldmaps} and all the satellite maps are from Google Map~\cite{googlemap}. 
We collect 651 pairs of historic maps and today's satellite images from 29 countries and show the distribution across countries in Figure~\ref{fig:appendix_map_distribution}. 
United States and Europe have the most historic maps online. 
For example, we obtain 194 images for the United States. 
In contrast, collecting maps from other regions is more difficult. For instance, we can only find 2 images for Australia and 4 images for Jersey.

\noindent \textbf{Task III: Mechanical Drawing Retrieval}. 
We collect 565 pairs of mechanical drawings and 3D rendered images from TraceParts~\cite{traceparts} and the other 589 pairs come from GrabCAD~\cite{grabcad}. These cover various specialized 3D components including brackets, nuts, gears, hinges, and clamps.

\begin{figure*}[t!]
\centering
\includegraphics[width=1\linewidth]{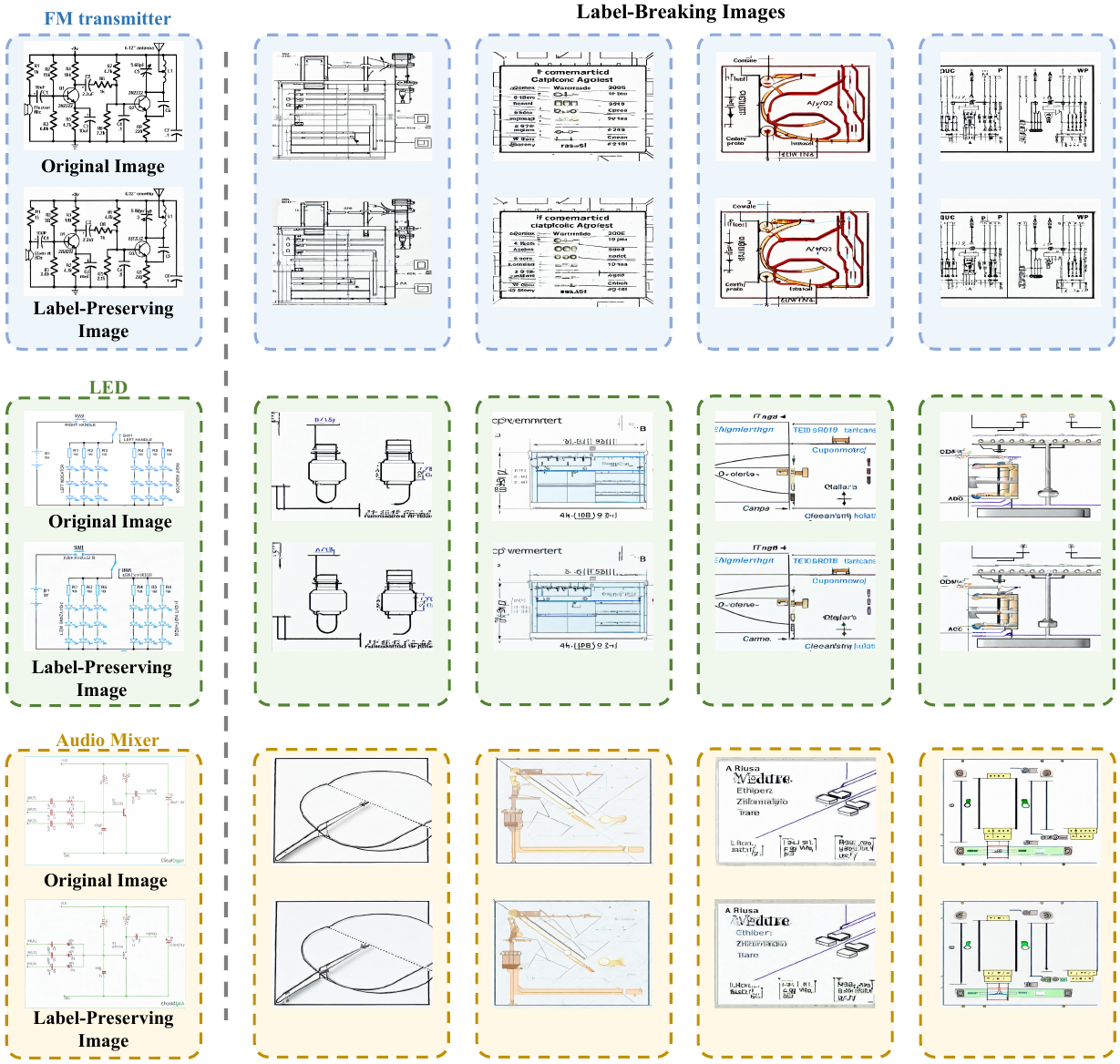}
\vspace{-1.5em}
\caption{\textbf{Generated Data of Circuit Diagrams}. We show the generated data of three circuit diagram images with different colors. In the left column, we present the original images and their label-preserving augmentations. In the other columns, we show the label-breaking images and the positive pairs for contrastive learning. For the images in each color box, there are minor differences between each other (zoom in to observe the differences), while the label-breaking images are totally different from their original images and have large variations. 
}
 \vspace{-1em}
\label{fig:appendix_generated_data_circuits}
\end{figure*}

\begin{figure*}[t!]
\centering
\includegraphics[width=1\linewidth]{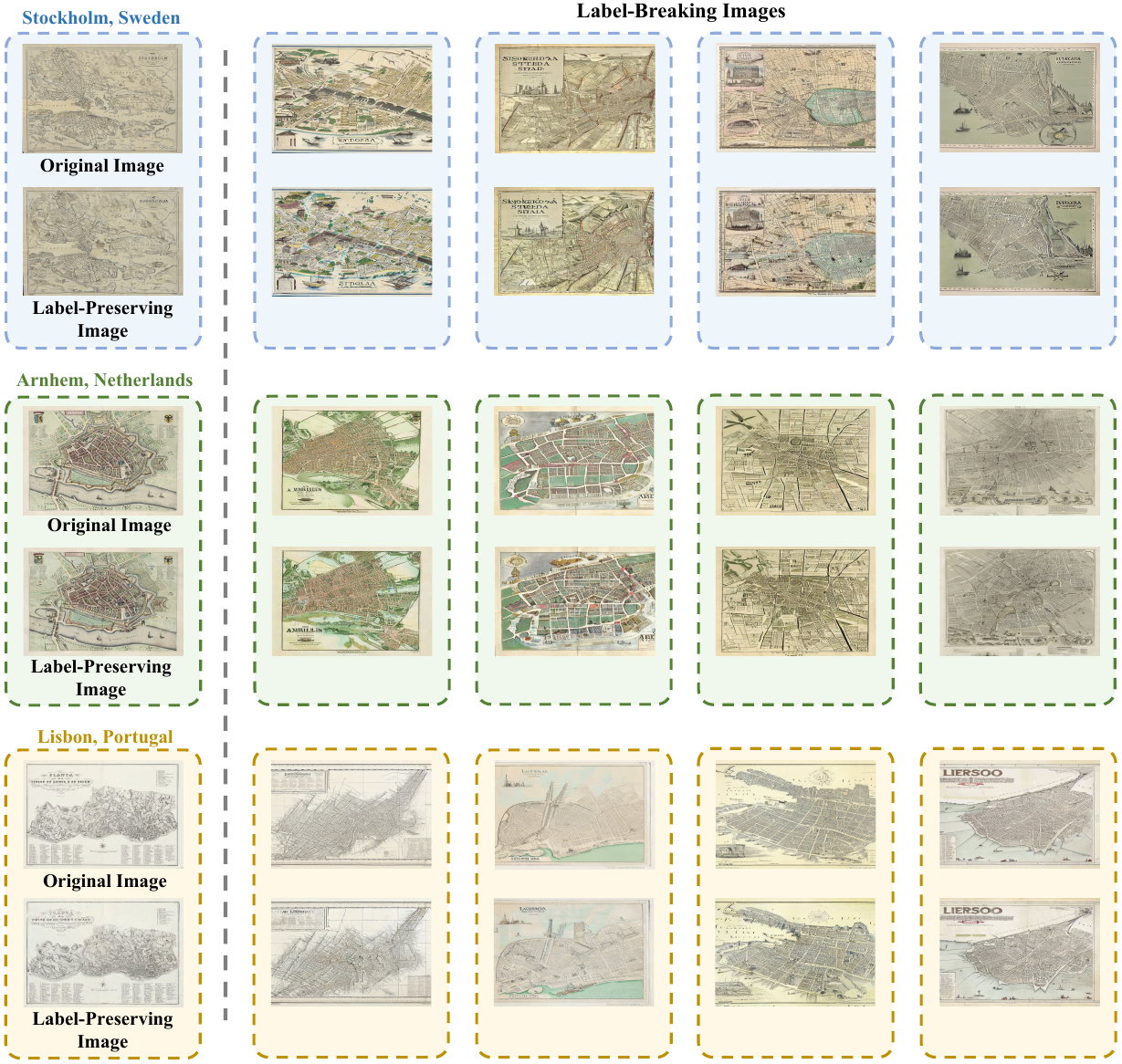}
\vspace{-1.5em}
\caption{\textbf{Generated Data of Historic Maps}. We show the generated data of three historic maps with different colors. In the left column, we present the original historic map with their label-preserving augmentations. In the other columns, we show the label-breaking images as well as the positive pairs for contrastive learning. For the images in each color box, there are minor differences between each other (zoom in to observe the differences), while the label-breaking images are totally different from their original images and have large variations. 
}
\label{fig:appendix_generated_data_maps}
\end{figure*}

\begin{figure*}[t!]
\centering
\includegraphics[width=1\linewidth]{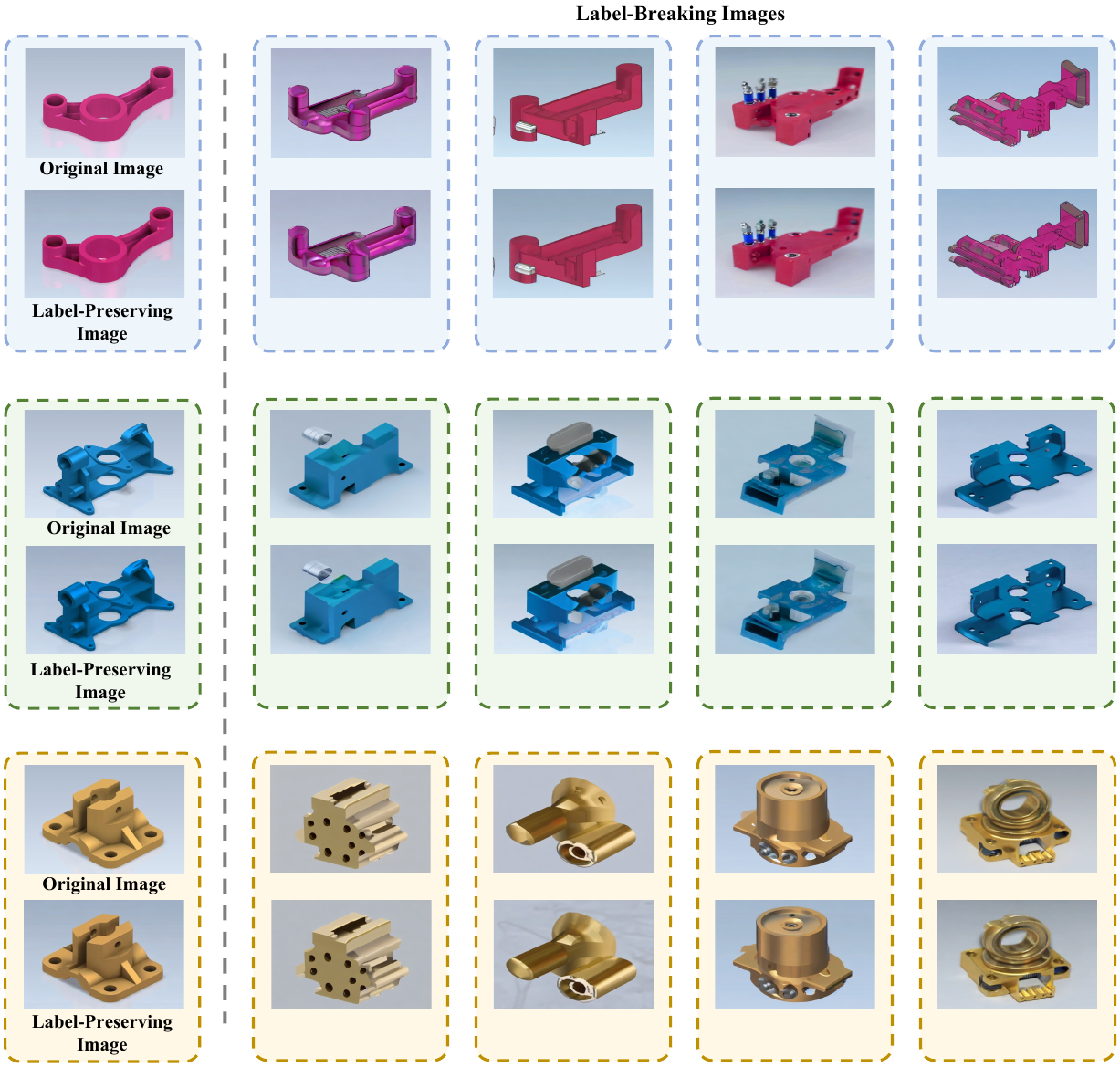}
\vspace{-1.5em}
\caption{\textbf{Generated Data of 3D Rendered Images} on Mechanical Drawing Retrieval. We show the generated data of three 3D rendered images with different colors. In the left column, we present the original 3D rendered images with their label-preserving augmentations. In the other columns, we show the label-breaking images as well as the positive pairs for contrastive learning. For the images in each color box, there are minor differences between each other (zoom in to observe the differences), while the label-breaking images are totally different from their original images and have large variations. 
}
\label{fig:appendix_generated_data_3d_rendered}
\end{figure*}

\begin{figure*}[t!]
\centering
\includegraphics[width=1\linewidth]{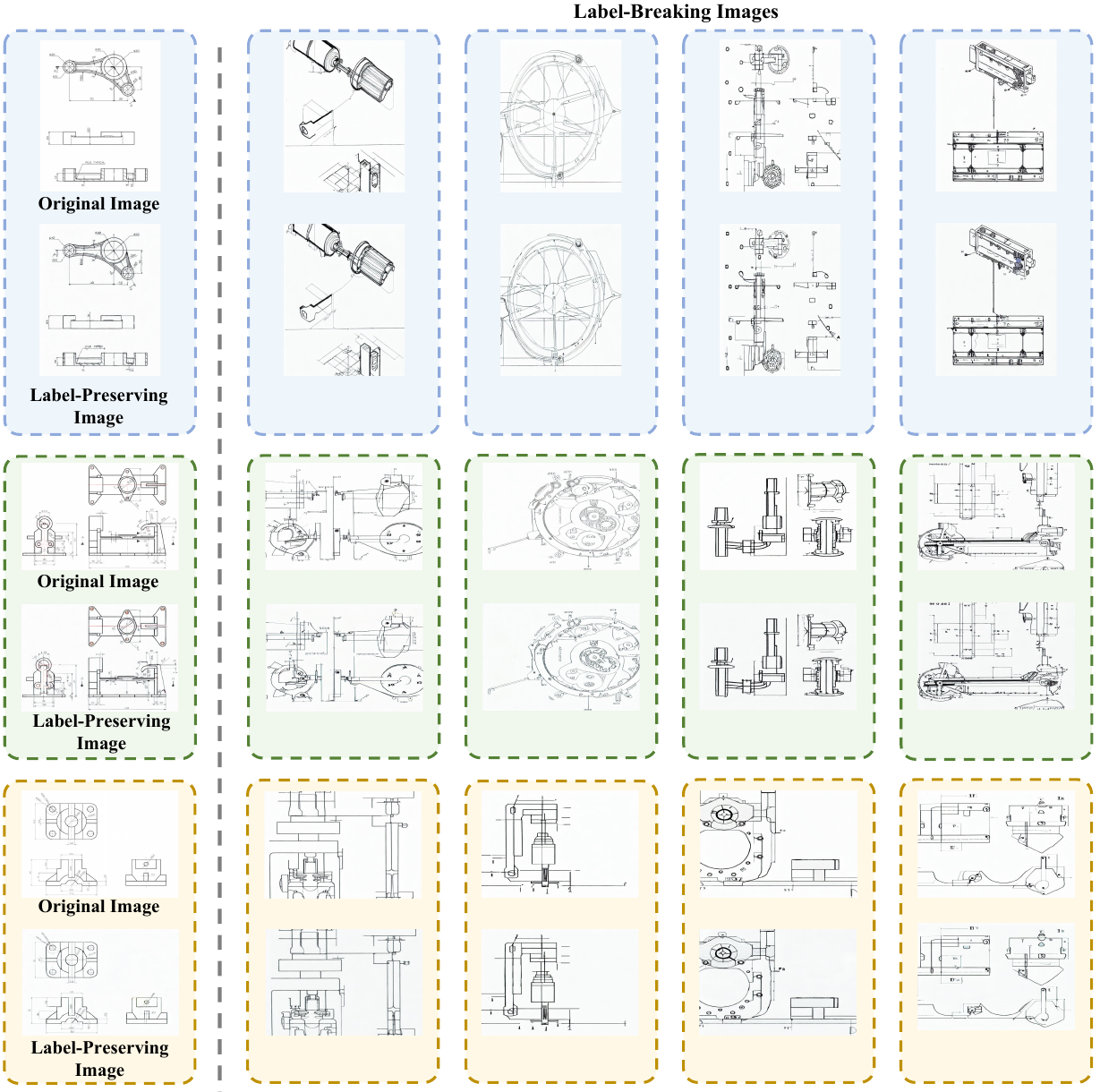}
\vspace{-1.5em}
\caption{\textbf{Generated Data of Mechanical Drawings}. We show the generated data of three mechanical drawings with different colors. In the left column, we present the original mechanical drawings with their label-preserving augmentations. In the other columns, we show the label-breaking images as well as the positive pairs for contrastive learning. For the images in each color box, there are minor differences between each other (zoom in to observe the differences), while the label-breaking images are totally different from their original images and have large variations. 
}
\label{fig:appendix_generated_data_mechanical}
\end{figure*}

\section{Visualizations}

\label{sec:appendix_visualization}

\noindent \textbf{Attention Maps from Vision Foundation Model} on low-resource images. We visualize the attention maps from vision foundation models without our low-resource baselines in Figure~\ref{fig:appendix_att_foundation}. This uses the ImageBind model adapted to our low-resource tasks with AdaptFormer. 
We show the highest activation on each region across all the attention maps of the middle transformer block. 
We can observe from the figure that only a few regions are activated for the three images. 
This means that the vision foundation model fails to understand the interaction between different image regions which is key for these specialized domains.   
As a result, the model cannot perform well on low-resource tasks. 
Thus, proper adaptation is needed for vision foundation models.

\noindent \textbf{Attention for Specialized Domains}. 
In Section~\ref{sec:adaptive}, we introduced our attention for specialized domains. Here, we visualize these learned attentions in Figure~\ref{fig:appendix_att_visualization}. Each domain tends to have a particular attention pattern with the different attention maps having different sizes of `receptive field' so that various levels of features can be encoded.
We observe that the attention for circuit diagram classification and mechanical drawing retrieval focuses more on vertical and horizontal regions. This is because these images have a lot of straight lines and right angles, which contain useful information. 
The attention maps for historic map retrieval highlight different local regions and tend to have much larger `receptive fields' than the other two tasks.

\noindent \textbf{Confusion Matrix}. 
We provide a confusion matrix in Figure~\ref{fig:supp_confusion}. 
Indeed, our baselines recognize components with prominent patterns more clearly, achieving stronger performance on \textit{dice}, \textit{infrared sensor}, and \textit{relay}. 
It confuses classes with shared components, \eg, \textit{audio mixer} and \textit{wave generator}.

\noindent \textbf{Label-Preserving and Label-Breaking Images}. In Section~\ref{sec:diverse}, we introduce our generated data for data scarcity. Here, we visualize the label-preserving and label-breaking images in Figure~\ref{fig:appendix_generated_data_circuits}, Figure~\ref{fig:appendix_generated_data_maps}, Figure~\ref{fig:appendix_generated_data_3d_rendered} and Figure~\ref{fig:appendix_generated_data_mechanical}. For label-preserving augmentations, there are only minor differences from their original images (zoom in to observe the differences). Thus, we use the original labels for them in model learning. However, for label-breaking augmentations, they are totally different from the original images. Therefore, we apply a self-supervised contrastive learning objective to learn from such data. We also present the augmentations for label-breaking images, which construct the positive pairs in contrastive learning. Note that each box in the figures denote a positive pair.


\end{document}